\documentclass[fleqn,10pt]{SelfArx} 

\usepackage{lipsum}

\usepackage{csvsimple}
\usepackage{booktabs}
\usepackage{longtable}
\usepackage{array}
\usepackage{siunitx}
\usepackage{pdflscape}
\usepackage{multirow}
\usepackage{graphicx}
\usepackage{caption}
\usepackage{subcaption}

\definecolor{color1}{RGB}{0,0,90} 
\definecolor{color2}{RGB}{0,20,20} 

\usepackage{hyperref} 
\usepackage{longtable}
\hypersetup{hidelinks,colorlinks,breaklinks=true,urlcolor=color2,citecolor=color1,linkcolor=color1,bookmarksopen=false,pdftitle={Title},pdfauthor={Author}}

\JournalInfo{Draft Version} 
\Archive{} 

\PaperTitle{A Comprehensive Machine Learning Framework for Micromobility Demand Prediction}

\Authors{Omri Porat\textsuperscript{1}*, Eran Ben-Elia\textsuperscript{2}, and Michael Fire\textsuperscript{3}} 
\affiliation{\textsuperscript{1}\textit{Department of Information Systems Engineering, Ben-Gurion University of the Negev, Beer Sheva, Israel}} 
\affiliation{\textsuperscript{2}\textit{Department of Information Systems Engineering, Ben-Gurion University of the Negev, Beer Sheva, Israel}} 
\affiliation{\textsuperscript{3}\textit{Department of Environmental, Geoinformatic \& Urban Planning, Ben-Gurion University of the Negev, Beer Sheva, Israel}} 
\affiliation{*\textbf{Corresponding authors}: omripor@post.bgu.ac.il, benelia@bgu.ac.il, mickyfi@bgu.ac.il}

\newcommand{\keywordname}{Keywords} 

\Abstract{ 
Dockless e-scooters, a key micromobility service, have emerged as eco-friendly and flexible urban transport alternatives. These services improve first- and last-mile connectivity, reduce congestion and emissions, and complement public transport for short-distance travel. However,  effective management of these services depends on accurate demand prediction, which is crucial for optimal fleet distribution and infrastructure planning.
While previous studies have focused on analyzing spatial or temporal factors in isolation, this study introduces a framework that integrates spatial, temporal, and network dependencies for improved micromobility demand forecasting. This integration enhances accuracy while providing deeper insights into urban micromobility usage patterns.
Our framework improves demand prediction accuracy by 27–49\% over baseline models, demonstrating its effectiveness in capturing micromobility demand patterns. These findings support data-driven micromobility management, enabling optimized fleet distribution, cost reduction, and sustainable urban planning.
}

\begin{document}


\maketitle 


\thispagestyle{empty} 


\section{Introduction} 
The growing adoption of micromobility, particularly 
e-scooters ~\cite{mcqueen2021transportation}, as a flexible~\cite{goh2022design}, and sustainable~\cite{fishman2021bikesharing, liu2022measuring} transportation solution is reshaping urban mobility~\cite{ABDULJABBAR2021102734}. 
E-scooters offer a promising solution for short-distance travel~\cite{abduljabbar2021role}. They can help reduce traffic congestion, lower emissions~\cite{moreau2020dockless}, and address first and last-mile connectivity challenges~\cite{zuniga2022evaluation}. In cities like Tel Aviv, e-scooters are integral in urban mobility~\cite{shahal2024towards}, reducing short car trips and complementing public transport ~\cite{esztergar2022assessment}.

Accurately predicting e-scooter demand remains a critical challenge~\cite{fearnley2020micromobility} for effective micromobility management~\cite{fearnley2020micromobility}. It is vital for fleet management, ensuring e-scooter availability~\cite{sanchez2022simulation}, and infrastructure planning~\cite{folco2023data}. Micromobility demand is highly dynamic and influenced by complex interactions among spatial, temporal, and network-related factors~\cite{castiglione2022delivering, hosseinzadeh2021factors}. Spatial factors~\cite{GEIPEL2024}, such as proximity to public transport hubs~\cite{LI2024104496, AARHAUG2023101279}, bike lanes~\cite{xu2024icn}, and land-use patterns~\cite{hosseinzadeh2021spatial,hong2023investigation}, significantly impact micromobility demand, while temporal factors such as weather conditions~\cite{ko2023analyzing, kimpton2022weather}, time of day~\cite{chiotti2023exploring}, and special events~\cite{boonjubut2022accuracy} influence demand. Additionally, the urban network structure~\cite{ham2021spatiotemporal, song2023sparse, arias2023exploring}, such as centrality~\cite{HU2024104477}, community structure~\cite{su14052564}, size, and density~\cite{freire2021unfolding} significantly influences e-scooter demand. Prior research enhances our understanding of the elements influencing e-scooter usage and offers valuable insights into the demand and factors affecting e-scooter usage.

Prior studies have employed time-series models, such as Prophet~\cite{Taylor2017},\footnote {Prophet~\cite{Taylor2017} - a time-series forecasting tool that handles seasonal patterns, missing data, and external regressors. } or LSTM-TPA (Temporal Pattern Attention)~\cite{Yu2024}, to forecast micromobility demand~\cite{jmse9111231, su14052564}. However, these models primarily capture temporal trends and fail to account for spatial dependencies.
 Similarly, spatial machine learning (GeoML) models struggle to capture temporal patterns~\cite{LI2022101848}. Moreover, GeoML and Time-Series models do not incorporate network structure~\cite{HOSSEINI2023101052}. 
 
 This study proposes a model that integrates spatial, temporal, and network-based features to improve predictive accuracy. We present a novel machine-learning framework that integrates spatial, temporal, and network features to enhance e-scooter demand prediction in urban settings.
(see Section~\ref{Results}).

\begin{figure*}[ht]\centering 
\includegraphics[width=\linewidth]{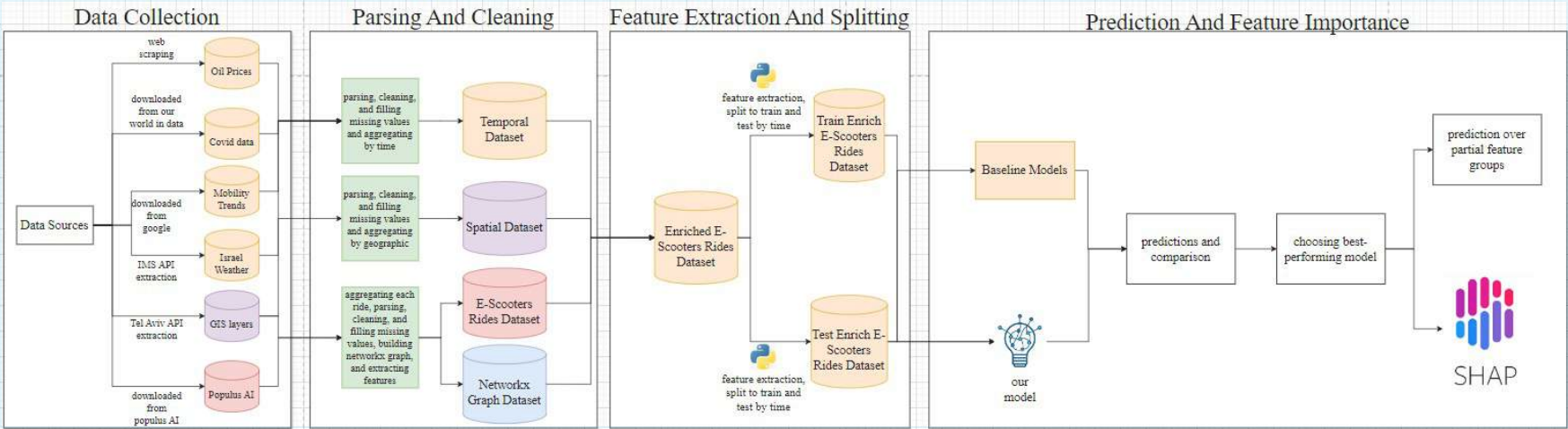}
\caption{Algorithm Workflow Overview}
\label{fig3:view}
\end{figure*}

Our methodology follows a four-phase process designed to adapt to diverse urban environments. The first phase involves data collection \textit{Data partitioning and preprocessing} across three main categories: (a) \textit{Temporal Information} - This includes standardized temporal data such as weather conditions, holiday schedules, public events, economic indicators, and mobility trends;
(b) \textit{Spatial information}- This category includes standardized GIS-based spatial data, such as public transport networks, land use patterns, and sociodemographic information;
(c) \textit{Network Information} - We gathered, standardized, and converted e-scooter usage data into a network format. In this representation, the nodes symbolize geographic areas, the edges depict the direct links between these areas, and the weights of the edges reflect the volume of e-scooter trips occurring between them.

The second phase involves extracting and processing features from each data category to prepare inputs for model training. The feature extraction process encompasses three main data categories.
Temporal feature extraction involves aggregating data into time-based features to capture seasonal patterns and trends in micromobility usage. Spatial feature extraction involves analyzing patterns across multiple levels—from macro-level (quarters) to micro-level (blocks)—by calculating metrics such as density and proximity using GIS layers. Network analysis involves representing e-scooter trips as a time-dependent directed graph. We analyze the e-scooter network on three levels: (1) Node level—measuring area connectivity using centrality metrics; (2) Edge level—analyzing trip patterns between geographic regions; (3) Graph level—assessing overall network connectivity. 

The third phase focuses on developing and evaluating a demand prediction model (see Section \ref{model_construction}) that integrates temporal, spatial, and network features. Our model combines a time series model to capture temporal patterns and a machine learning ensemble algorithm. We compare our algorithm to several baseline algorithms that use temporal or spatial features. For each algorithm, we evaluate performance using multiple metrics such as Mean Average Error (MAE) and Root Mean Squared Error (RMSE).

Lastly, to interpret our model predictions, we conducted  \textit{Feature Analysis} using Explainable AI techniques, such as SHapley Additive exPlanations (SHAP) \cite{lundberg2017unified}. 
SHAP values quantify each feature’s contribution to the model’s predictions, identifying key factors influencing e-scooter demand.
This approach provides insights into the contributions of spatial, temporal, and network features, improving interpretability and understanding of the model’s results .

We conducted a large-scale experiment to evaluate our proposed methodology. We used real-world e-scooter data from Tel Aviv provided by Populus AI. It contains approximately 16 million rides (April 2020 to February 2023). We integrated six additional data sources, such as GIS layers, Google Mobility Trends, Energy Prices, and Weather data. In total, we employed 341 features - 238 spatial features, 86 temporal features, and 17 network features.  (see Table \ref{Features_Table}). We aggregate the data across multiple spatial and temporal levels. We subsequently implemented a temporal cut-off method to avoid data leakage. We then used seven different prediction models for comparative analysis and found the best-performing model.


The fully integrated model demonstrated a 27–49\% reduction in MAE compared to baseline feature subsets (see Section~\ref{Results}), underscoring the predictive advantage of combining spatial, temporal, and network features. Notably, the inclusion of network centrality metrics contributed the most significant improvement in forecasting accuracy. Our analysis indicates that our model reduced Mean Absolute Error (MAE) by 66-91\% compared to the state-of-the-art baseline time series model (see Section~\ref{Results}). Furthermore, our model significantly outperformed baselines relying on individual feature types, reinforcing the importance of integrating spatial, temporal, and network features. These results confirm our model's ability to capture seasonal trends and improve the accuracy of micromobility demand predictions (refer to Section \ref{Results}). It demonstrates the significance of integrating diverse features to achieve precise demand forecasting for micromobility solutions.

We assessed the performance of our best-performing model by using only certain groups of features. There are varying degrees of improvement based on the specific types of features included in the evaluation. Specifically, our model, using temporal and network features, improved prediction accuracy by 2-21\% compared to network, spatial, and temporal features, 34-63\% compared to spatial and temporal features, and 19-39\% compared to network and temporal features.

Our results indicate that network features, particularly centrality measures, were the most influential predictors of micromobility patterns in our model (see section \ref{Results}). SHAP analysis identified node degree and path length as key predictors of e-scooter demand, indicating that areas with higher connectivity within the urban mobility network exhibit greater adoption of micromobility services. It suggests that well-integrated road networks play a vital role in shaping e-scooter usage patterns. These findings underscore the critical role that urban network structure plays in shaping e-scooter usage patterns.

The remainder of this paper is structured as follows: Section \ref{Related_Work} reviews related work on e-scooter ride patterns, feature selection, and machine learning approaches. Section \ref{Methodology} We \ref{Methodology}, and the experimental framework \ref{experiment_tel_aviv}. Section \ref{Results} presents results and performance comparisons with baseline models. Section \ref{Conclusion} discusses implications, limitations, and conclusions. Finally, Section \ref{Future_Work} outlines directions for future research.

\section{Related Work}
\label{Related_Work}
This section reviews prior research on micromobility demand prediction, focusing on spatial, temporal, and network-based factors, as well as machine learning approaches. First, we analyze temporal and spatial patterns affecting micromobility demand (see Section~\ref{Spatial, Temporal, and User-Related Factors in Micromobility Demand}), such as weather conditions, land use and connectivity. Second, we explore machine-learning approaches in micromobility, focusing on time-series models (Section~\ref{Machine Learning Approaches for Micromobility Demand Prediction}). Finally, we review SHAP and Prophet and their role in understanding micromobility behavior (Section~\ref{SHAP_and_Prophet}).

\subsection{Spatial, Temporal, and User-Related Factors in Micromobility Demand}
\label{Spatial, Temporal, and User-Related Factors in Micromobility Demand} 

Understanding e-scooter demand requires analyzing spatial and temporal factors that shape micromobility usage. Prior studies have examined these influences in diverse urban settings, providing insights into demand fluctuations \cite{shahal2024towards,hassam2024spatiotemporal,liu2019associations}.
\label{spatial_features}
Studies indicate that road infrastructure significantly influences e-scooter adoption, with key factors including cyclist comfort, road maintenance, and route flexibility. For example, Feng et al.~\cite{Feng2020} found a strong correlation between e-scooter adoption and cyclist comfort ratings in Austin, Texas, suggesting that well-maintained cycling infrastructure encourages usage. Similarly, Lazarous et al.~\cite{Lazarus2020} showed that dockless e-scooters outperformed e-bikes in less populated areas in San Francisco, demonstrating user preferences for more flexible transportation modes. Rodriguez et al.~\cite{rodriguez2022travel} showed that better-maintained road infrastructures in Puerto Rico significantly increase e-scooter adoption. These findings demonstrate the vital role of infrastructure in shaping micromobility usage patterns. 
User demographics and the interplay with other transportation modes are equally important. Liao and Correia, Sophia et al., Bretones and Orio, and Christoforou et al.~\cite{liao2022electric,sophia2021uses, bretones2022sociopsychological, christoforou2023neighborhood} found that e-scooter users were predominantly young, educated, and affluent. In central Austin, Caspi et al. In Taipei, Liu and Lin~\cite{LIU2022107} demonstrated similarities in travel time and distance but differences in high-demand areas and rush hour usage. Taipei, Liu and Lin~\cite{LIU2022107} demonstrated similarities in travel time and distance but differences in high-demand areas and rush hour usage. 

The presented studies discuss different transportation modes, user profiles, and road infrastructure as predictors for e-scooter usage. 
While these studies provide valuable insights, they often focus on spatial or temporal features, without integrating them. Moreover, they overlooked network features, which are vital in micromobility demand \cite{folco2023data}.

Temporal factors, such as weather and fuel prices, significantly affect micromobility demand. Mehzabin Tuli et al.~\cite{MEHZABINTULI2021164} developed a regression model incorporating these variables to predict e-scooter demand patterns. Among temporal factors, weather conditions, such as rainfall and temperature, significantly influence micromobility usage patterns. Noland~\cite{noland2021scootin} showed that rain and cold temperatures significantly influenced micromobility usage. Also, Xin et al.~\cite{xin2022impact} showed that the pandemic led to fewer trips and fewer rush-hour rides while long-duration trips increased. Integrating these diverse findings provides a comprehensive view of the main factors influencing micromobility usage patterns. 

While prior research has explored these factors separately, an integrated approach that considers spatial, temporal, and network features is essential for improving predictive performance.

\subsection{Machine Learning Approaches for Micromobility Demand Prediction}
\label{Machine Learning Approaches for Micromobility Demand Prediction}
Building on these demand-influencing factors, researchers have explored machine-learning techniques for micromobility prediction, ranging from classical regression models to deep-learning approaches.

\subsubsection{Classical Machine Learning Approaches}
Several studies have applied traditional machine learning techniques, including regression~\cite{chiotti2023exploring, boonjubut2022accuracy} and ensemble models~\cite{ko2023analyzing, peng2022assessing}, to forecast micromobility demand. These models often capture correlations between external features, such as weather, land use, sociodemographic features, and micromobility trip volumes. 
Early studies applied regression models to predict micromobility demand based on spatial or temporal features. For instance, Wang et al.~\cite{wang2020traffic} applied Random Forest (RF) and Gradient Boosting Machines (GBM) to predict e-scooter demand in Beijing. Additionally, Maren Schieder~\cite{schnieder2023ebike} demonstrated the effectiveness of RF models for bike-sharing demand prediction in London.

More recent studies, such as Zuniga-Garcia et al.~\cite{Zuniga-Garcia2022}, applied XGBoost to predict e-scooter demand in Austin, Texas and demonstrated enhanced accuracy compared to linear models in capturing non-linear dependencies. Lin et al.\cite{lin2024insights} developed an adaptive XGBoost model for shared bicycle demand prediction in Shanghai, combining both weather and urban morphology features. 

Despite their advantages, classical ML approaches assume feature independence~\cite{hall1999correlation}, which is problematic for micromobility demand data~\cite{xu2024icn, xu2023real} due to inherent spatial~\cite{shin2022influence} and temporal~\cite{zhao2021impact} autocorrelations. Short-term demand fluctuations and seasonal effects which necessitate more sophisticated modeling techniques. 

\subsubsection{Temporal Machine Learning Approaches}
\label{Temporal_ML_approaches}
Due to the seasonal and short-term fluctuations in micromobility demand, researchers have explored time-series forecasting techniques for improved prediction. Some of the methods include Prophet~\cite{Taylor2017}, LSTMs~\cite{greff2016lstm}, and GRU-based (Gated Recurrent Units) architectures~\cite{dey2017gate}.

Wang et al.~\cite{jmse9111231} applied Prophet, an open-source tool used for forecasting time series data, to predict micromobility usage in New York City, achieving enhanced results than traditional time series approaches by combining holidays, events, weather conditions, and mobility trends. However, Prophet struggles to model non-linear interactions and lacks the ability to capture spatial dependencies in micromobility demand.

Deep learning approaches, mainly LSTM (Long Short-Term Memory) networks, capture long-term and short-term seasonality. For example, Yu et al.~\cite{Yu2024} developed an LSTM-TPA (Temporal Pattern Attention) model to forecast e-scooter demand in Chicago, demonstrating its ability to capture short-term fluctuations during special events and holidays. The LSTM provided superior accuracy to traditional ML models by learning complex temporal dependencies. 

While time-series models (Prophet, LSTM, GRU) effectively capture temporal dependencies, they fail to incorporate spatial relationships and network structures—key elements influencing micromobility demand. 

Despite advancements in spatial and temporal micromobility forecasting, existing models overlook the impact of network connectivity on demand patterns. This study addresses this gap by integrating spatial, temporal, and network-based features into a unified machine-learning framework.

\subsection{Time Series and Model Explainability Tools}
\label{SHAP_and_Prophet}

This study utilizes Prophet~\cite{Taylor2017} for time-series forecasting and SHAP~\cite{lundberg2017unified} for model explainability. Prophet captures seasonal trends in micromobility demand, while SHAP quantifies the influence of individual features on model predictions.

Prophet~\cite{Taylor2017} is a general-purpose time-series forecasting model for micromobility demand prediction. Wang et al.~\cite{wang2020traffic} used Prophet to forecast e-scooter demand in New York City, integrating weather conditions and public holidays as key predictors.
Similarly, Hernandez et al. \cite{HERNANDEZ2024101015} used Prophet to analyze micromobility seasonality in Chicago, capturing daily commuting patterns and weekend usage distributions.

SHAP (SHapley Additive Explanations)~\cite{lundberg2017unified} enhances model interpretability by assigning importance scores to individual features, explaining their contribution to predictions. Shalit et al. \cite{shalit2023supervised} used SHAP to identify prominent features imputing missing boarding stops in smart card data. In micromobility research, Zuniga-Garcia et al.~\cite{Zuniga-Garcia2022} applied SHAP to analyze e-scooter demand in Austin. Their findings highlighted bike lane density and public transport proximity as primary influencing factors. Similarly, Shahal et al.~\cite{SHAHAL2024722} employed SHAP to quantify weather-related influences on e-scooter demand in Tel Aviv, reinforcing the role of environmental factors in micromobility forecasting.

\section{Methods}
\label{Methodology}
This study proposes a machine-learning framework to predict micromobility demand across different geographic scales. The methodology follows a two-stage approach: (1) model development and (2) experimental evaluation. We develop a generalized machine-learning model for micromobility demand prediction, ensuring adaptability across different urban environments. Our framework consists of four key steps: (1) data preprocessing, (2) feature engineering, (3) model construction, and (4) model evaluation. Each step is detailed in the following subsections.

\subsection{Data Preprocessing}
\label{Data_Processing}
Micromobility ride data is collected from municipal agencies, service providers, and open data portals. Each record includes trip origin, destination, start and end times, and ride duration. We first clean the data by removing trips with empty fields and removing abnormal trips, which are trips with a duration of less than $t_{min}$ seconds or higher than $t_{max}$.

We aggregate the ride data based on a temporal variable, $T_{scale}$. The variable represents different time scales, such as hour, day, or month. This temporal aggregation allows for the examination of trends at varying levels of granularity.
That allows us to analyze short-term fluctuations and long-term trends in micromobility demand. 

Spatial aggregation $S_{level}$ is performed at multiple levels, ranging from large city quarters to fine-grained hexagonal grid units.
The multi-scaled approach guarantees that predictions can adjust to macro and micro changes and different urban planning requirements. 
The micromobility dataset is enriched with external data sources to incorporate broader urban context. These sources include:

\textbf{Temporal Data.} We include weather conditions, such as temperature, wind speed, humidity, and precipitation, as these factors affect micromobility demand. We combine public events and holidays, which can affect micromobility demand. 
Then, we combine economic indicators, such as daily energy prices, to assess their influence on micromobility adoption. We then add mobility trends, such as residential and transport time, to evaluate their effect on micromobility demand fluctuations.   

\textbf{Spatial Data.} Our framework integrates spatial features including urban infrastructure and road networks. These features are extracted from points of interest, public transport hubs, and land use classifications to predict micromobility demand patterns.
The list of spatial features with their descriptions is in Table \ref{Spatial_Features_Table}. We categorized the spatial features into three geometric types based on their shape:
Point Features ($P$): Elements that represent discrete locations in space, defined by coordinate pairs $(x,y)$. These include specific points of interest, such as docking and bus stations.
Line Features ($L$): Elements that represent vectors between two or more points, defined by a series of connected coordinates ${(x_1,y_1),(x_2,y_2),...,(x_n,y_n)}$. These features capture paths, such as metro lines and bike lanes.
Polygon Features ($A$): Elements characterized by closed boundaries that enclose specific areas, represented as a sequence of coordinates, denoted as ${[(x_1,y_1), (x_2,y_2),...,(x_n,y_n),(x_1,y_1)]}$.\footnote{Polygon is a closed geometric shape formed by connecting a sequence of points (vertices) in a plane. It is mathematically represented as ${[(x_1,y_1), (x_2,y_2),...,(x_n,y_n),(x_1,y_1)]}$, where each $(x_i,y_i)$ represents the coordinates of a vertex, with the first point repeated at the end to close the shape.} These features represent land use zones and districts, such as parks, commercial areas, and residential zones.

\textbf{Network Data.} We model the micromobility ride network as a time-dependent directed graph $G_{t}^{s} = (V_{t}^{s}, E_{t}^{s}, W_{t}^{s}) $, where $V_{st}$ represents nodes (geographic areas) at time $t$ and spatial division $s$, and edges $E_{t}^{s}$ capture ride connections weighted by trip volume $W_{t}^{s}$. Each node represents a geographic area. $E_{t}^{s}$ is the set of directed edges at time $t$, and spatial division $s$, where an edge $(v_{i}^{s}, v_{j}^{s})$ exists if there are recorded micromobility trips from area $v_{i}^{s}$ to $v_{j}^{s}$. $W_{t}^{s}$ is the set of edge weights at time $t$ and spatial division $s$, $w_{ij}^{s}$ represents the volume of trips between areas $v_{i}^{s}$ to $v_{j}^{s}$.  The network uses origin-destination pairs from the micromobility ride data, providing insight into connectivity patterns and centrality. We join the data sources with micromobility rides data. We then perform data normalization using min-max scaling and filling in missing values to ensure we can perform feature extraction and model training over the data.

\subsection{Feature Engineering}
\label{Feature_Engineering}
In this study, we predict the micromobility flow between spatial regions at a given time, represented as the weight of an edge in the time-dependent network. We integrate the micromobility rides dataset with the temporal, spatial, and network attributes to predict these flows. We categorize and compute the extracted features into spatial, temporal, and network features: 

\textbf{Temporal Data (see Table~\ref{Temporal_Features_Table_Tel_Aviv})} We integrate for each $T_{scale}$ and $S_{level}$ temporal datasets to enhance micromobility demand predictions. 
Data acquisition from open APIs and government portals enables the extraction of key features, which are categorized as follows:

\begin{itemize}
\item \textit{Mobility Trends:} We gather historical data from Google Mobility Trends to analyze how much time people spend in different locations, including workplaces, residences, transportation, and recreational areas. We use mobility data that precedes our prediction periods to avoid data leakage. These insights help us understand the overall patterns and dynamics of transportation.

\item \textit{Special Events:} We extract features such as weekdays /weekends and holidays. Those features can capture temporary demand hotspots.

\item \textit{Energy Prices:} We extract features, including WTI and Brent oil prices, which precede our prediction periods. These prices have the potential to influence the overall demand for micromobility services.

\item \textit{Weather Conditions:} We extract features such as temperature, wind speed, rain, and humidity forecast for our prediction periods. Those features can affect overall micromobility demand.

\item \textit{Seasonality Patterns:} We include seasonality-based features from Prophet to enhance prediction accuracy that capture periodic demand fluctuations, such as holiday effects. It captures patterns from our temporal data and adds to the final dataset.
\end{itemize}

\textbf{Spatial Features  (see Table \ref{Spatial_Features_Table}): } We divide the GIS layers into three primary feature types: Point-based, Line-based, and Polygon-based features. Each category provides spatial attributes that influence micromobility patterns. The extracted features are detailed as follows:

\begin{itemize}

\item \textit{Point-Based Features:} We use point-based features such as e-bike stations, e-scooter docking points, public transportation stops, train stations, commercial stores, schools, hospitals and government buildings. We determine numerical indicators by calculating the number of facilities within each designated spatial unit for each feature. For instance, if a spatial unit contains three bus stations, we classify it under the feature "number of bus stations." This count-based approach allows us to capture the density and distribution of urban infrastructure within each analysis zone.

\item \textit{Line-Based Features:} We use line-based features such as bike lanes, pedestrian paths, highways, main roads, metro lines and bus lanes. We calculate the total length of relevant road segments within each spatial unit for each feature. 

\item \textit{Polygon-Based Features:} We use polygon-based features such as parks, nature reserves, housing, and commercial areas. We calculate the total area of different land-use categories within each spatial unit for each feature. 

\end{itemize}

\textbf{Network Features (see Table~\ref{Network_Features_Table}): } 
For each $T_{scale}$ and $S_{level}$, we extract from the time-dependent directed graph $G_{t}^{s}$ network features and divide them into three primary feature types: Node-based, Edge-based, and Network-based. Each category provides attributes that influence micromobility patterns.
The extracted features are detailed as follows (see Table~\ref{Network_Features_Table}):

\begin{itemize}
\item \textit{Node-Based Features:} We use node-based features such as betweenness centrality~\cite{brandes2001faster}\footnote {Betweenness centrality of a node is the sum of the fraction of all-pairs shortest paths that pass through the node} and degree centrality~\cite{berman1994nonnegative}.\footnote{The degree centrality for a node v is the fraction of nodes connected to.} We calculate it for each node in the graph.

\item \textit{Edge-Based Features:} We use edge-based features such as edge betweenness centrality~\cite{brandes2001faster}\footnote {Betweenness centrality of an edge is the sum of the fraction of all-pairs shortest paths that pass through the edge~\cite{brandes2001faster}.}  and edge connectivity~\cite{esfahanian2013connectivity}\footnote {The edge connectivity is equal to the minimum number of edges that must be removed to disconnect G or render it trivial. If source and target nodes are provided, this function returns the local edge connectivity: the minimum number of edges that must be removed to break all paths from source to target in G~\cite{esfahanian2013connectivity}.}  We calculate it for each edge in the network.

\item \textit{Network-Based Features:} We use network-based features such as average degree connectivity~\cite{barrat2004architecture}\footnote {Compute the average degree connectivity of graph.
The average degree connectivity is the average nearest neighbor degree of nodes with degree k~\cite{barrat2004architecture}.} and clustering coefficients~\cite{fagiolo2007clustering}.\footnote {For unweighted graphs, the clustering of a node is the fraction of possible triangles through that node that exist~\cite{fagiolo2007clustering}.} We calculate it for the whole graph.

\end{itemize}

\subsection{Constructing Prediction Model} 
\label{model_construction}
In this section, we construct a micromobility demand prediction framework based on the extracted features described in Section~\ref{Feature_Engineering}. Our approach combines regression-based models with time-series forecasting to enhance prediction accuracy. We used seven machine-learning models to evaluate our model: Decision Tree~\cite{breiman2017classification}, Elastic Net~\cite{zou2005regularization}, Lasso Regression~\cite{tibshirani1996regression}, Ridge Regression~\cite{hoerl1970ridge}, Random Forest \cite{breiman2001random},
\footnote{We ran the models with default values. Then, we ran XGBoost and Random forest with $n_{estimators}$ of 1000} Nearest Neighbors~\cite{cover1967nearest}, and Prophet~\cite{Taylor2017}. Prophet is a time-series forecasting model designed to capture seasonality, short-term fluctuations, and long-term trends.

Our prediction framework consists of three main steps: data splitting, model evaluation, and feature importance analysis.
In data splitting, for each $T_{scale}$ and $S_{level}$, we apply temporal cut-off\footnote{Temporal Cut-Off is a point in time used to divide a dataset into training and testing sets} to ensure that our evaluation reflects real-world forecasting and prevents data leakage. 
In model evaluation, we train and evaluate the performance of seven machine learning models using evaluation metrics such as Mean Absolute Error (MAE), Root Mean Squared Error (RMSE), and Mean Absolute Percentage Error (MAPE).

\subsection{Feature Group Analysis}
\label{Model_Evaluation}

After evaluating the seven predictions models with all the features, we then select the best-performing model. Next, we examine the contribution of different feature groups to the model performance. 

Additionally, we train and evaluate the best-performing model using three distinct feature groups- spatial features, network features, and temporal features and combine them: 

\begin{itemize}
\item \textit{Network and temporal features (see Tables~\ref{Network_Features_Table} and~\ref{Temporal_Features_Table_Tel_Aviv})} - Includes network characteristics, such as connectivity and centrality, with temporal attributes, such as mobility trends, energy prices, and holidays.
\item \textit{Spatial and temporal features (see Tables~\ref{Spatial_Features_Table} and~\ref{Temporal_Features_Table_Tel_Aviv})} - Includes spatial characteristics, such as public transportation routes, land use, and points of interest, with temporal attributes, such as mobility trends, energy prices, and holidays.
\item \textit{Network and spatial features (see Tables~\ref{Spatial_Features_Table} and~\ref{Network_Features_Table})}-  Includes network characteristics, such as connectivity and centrality, with spatial characteristics, such as public transportation routes, land use, and points of interest.
\end{itemize}

Finally, we conduct a feature importance analysis for the best-performing model using SHAP values to identify the most influential features for micromobility demand prediction. The insights can support policymakers in optimizing urban mobility by identifying features and feature groups that influence on micromobility adoption.

\section{Experiments}
\label{experiment_tel_aviv}
We test the effectiveness of our methodology using real-world e-scooter data from Tel Aviv. 
This phase involved structured experimentation to assess model performance in a real urban environment.

\subsection{Datasets and Data Processing}
\label{Datasets}
\textbf{Datasets.}
We integrated six datasets, each adding crucial spatial, temporal, or network-based attributes to strengthen our micromobility demand predictions model: 
\begin{itemize}
\item \textit{E-Scooter Ride Data (Populus AI)~\cite{populusAIRidesData}} A dataset that comprised approximately 16 million e-scooter trips recorded in Tel Aviv from April 2020 to February 2023, including trip duration, distance, start and end times, and routes. 

\item \textit{Tel Aviv GIS Data~\cite{tel_aviv_open_data}}– A comprehensive dataset that integrates 235 spatial layers of urban infrastructure, public transport centers, and cycling pathways.
\item \textit{COVID-19 Data (Our World in Data)~\cite{owid-coronavirus}} – Global pandemic-related metrics affecting urban mobility. The dataset contains 98 features.
\item \textit{Fuel Price Data~\cite{ourworldindata_crude_oil_prices}}– Weekly prices of global oil and gas from 2020 to 2023. The dataset contains two features, WTI prices and Brent prices.
\item \textit{Google Mobility Trends~\cite{owid-covid-mobility-trends}} – City-level movement trends across different sectors. The dataset contains 2 features.
\item \textit{Weather Data~\cite{israeli_weather}} – Daily temperature, precipitation, and wind speed records from the Israel Meteorological Institute.
\end{itemize}

After integrating the data from the six datasets, We first cleaned the e-scooter ride dataset by handling missing values and removing trips with a duration of less than 30 seconds or higher than 2 hours.
We chose those trips duration to eliminate outliers or system malfunction. The average micromobility trip in Tel Aviv is between 10-15 minutes.

\subsection{Feature Extraction}
\label{Data_Processing_Feature_Extraction}
After integrating the data from the six datasets, We first cleaned the e-scooter ride dataset by handling missing values and removing trips with a duration of less than 30 seconds or higher than 2 hours.

Next, we extracted features by aggregating data across temporal and spatial dimensions to ensure a structured and meaningful representation. 
We aggregated the data across the three $T_{scale}$: monthly, daily, and hourly, to capture long-term trends and short-term fluctuations. Later, we performed spatial aggregation over six $S_{level}$: quarters, subquarters, neighborhoods, city blocks, statistical areas, and hexagonal grids. We have 18 combinations of spatial and geographic levels in total. We categorized the extracted features into three primary types: temporal, spatial, and network features.

\textbf{Temporal Features (see Table~\ref{Temporal_Features_Table_Tel_Aviv})}
For each $T_{scale}$ and $S_{level}$, we integrate the external datasets to enhance micromobility demand prediction. The temporal features include:

\begin{itemize}
\item \textit{Mobility Trends:} Features that measure fluctuations in visits to different locations in Tel Aviv, including workplaces, residences, transportation hubs, and park areas. 
\item \textit{Special Events:} We extract features such as weekdays / weekends and Jewish holidays.
\item \textit{Energy Prices:} We extract features, including WTI and Brent oil prices in Israel.
\item \textit{Weather Conditions:} We extract weather features in Tel Aviv, such as temperature, wind speed, rain, and humidity.
\item \textit{Seasonality Patterns:} We include seasonality-based features from Prophet to enhance prediction accuracy that capture periodic demand fluctuations, such as Jewish holiday effects and weekdays prior scale. 
\end{itemize}

The final dataset includes 86 temporal features, all detailed in Table \ref{Temporal_Features_Table_Tel_Aviv}.

\textbf{Spatial Features (see Table~\ref{Spatial_Features_Table_Tel_Aviv})}
For each $T_{scale}$ and $S_{level}$, we extract spatial features from the GIS layers by dividing the data into three feature types: point-based features - $P$, line-based features - $L$, and polygon-based features.
\begin{itemize}
\item \textit{Point-Based Features:} We 135 use point-based features such as The number of Tel-O-Fun docking stations, Dan and Egged bus stations, Kindergartens, and Israel Rail stations, within each $S_{level}$.
\item \textit{Line-Based Features:} We use 31 line-based features such as the length of Tel-O-Fun lanes, highways, M1-M3 light train (metro) lines and bus lanes within each $S_{level}$.
\item \textit{Polygon-Based Features:} We use 72 polygon features such as the total area of parks and commercial areas within each $S_{level}$.
\end{itemize}
We extract 238 spatial features, provided in Table \ref{Spatial_Features_Table}.

\textbf{Network Features (see Table~\ref{Network_Features_Table})}
we construct the time-dependent directed network $G_{t}^{s}$.\footnote{We used NetworkX~\cite{hagberg2008exploring} - a Python package for the creation, manipulation, and study of the structure, dynamics, and functions of complex networks. Nodes represented geographic areas within specific time frames, while edges represented the number of trips between nodes, with weights capturing trip volumes}
We extracted 17 network features, categorized as follows:
\begin{itemize}
\item \textit{Node-Based Features:} We use node-based features such as betweenness centrality and calculate it for each node in the network, such as, betweenness centrality of Yad Eliyhau Neighborhood.
\item \textit{Edge-Based Features:} We use edge-based features such as edge betweenness centrality and edge connectivity and calculate it for each edge in the network, such as the connectivity between the the Ramat Aviv and Nave Shaan neighborhoods.
\item \textit{Network-Based Features:} We use edge-based features such as average degree connectivity and clustering coefficients and calculate it for the whole network.
\end{itemize}


In total, we used 341 features- 238 spatial features, 86 temporal features, and 17 network features.

\subsection{Model Evaluation}
We split the dataset into training (2020–2022) and testing (2022–2023) subsets using a temporal cut-off approach, as described in Section~\ref{model_construction}.

We trained the best-performing model in three distinct feature sets:
\begin{itemize}
\item \textit{Network and Temporal Features}: provided in Tables~\ref{Temporal_Features_Table_Tel_Aviv} and~\ref{Network_Features_Table}.
\item \textit{Spatial and Temporal Features}: provided in Tables~\ref{Spatial_Features_Table_Tel_Aviv} and~\ref{Temporal_Features_Table_Tel_Aviv}.
\item \textit{Network and Spatial Features}: provided in Tables~\ref{Spatial_Features_Table_Tel_Aviv} and~\ref{Network_Features_Table}.
\end{itemize}

Lastly, we applied SHAP-based feature importance analysis to interpret the key factors driving micromobility demand (see Figure~\ref{fig3:view}).

\section{Results} 
\label{Results} 

In this section, we present the experiment results. First, we present the e-scooter ride data and network characteristics with temporal and spatial analysis. Second, we present our model results and best-performing model results over partial feature groups and SHAP analysis results.   
The e-scooter rides dataset includes approximately 16 million rides, with an average trip duration of 14 minutes and an average distance of 2.5 kilometers per trip, from April 2020 to February 2023. After removing incomplete records and trips shorter than 30 seconds or longer than 2 hours, the data set remains with 15.4 million rides.

We split the dataset into training (April 2020 - October 2022) and testing (November 2022 - February 2023) using a temporal cut-off approach. The train data contains 13.8 million trips, while the test data contains 1.6 million. 

Then, we constructed a time-dependent directed network $G_{t}^{s}$, for each $T_{scale}$ and $S_{level}$, where the network structure is $T_{scale}$ invariant. The following table describe the network property for each $S_{level}$ (see Table~\ref{Network_Stats}).

\renewcommand{\thetable}{S\arabic{table}}
\begin{table}[h]
    \centering
        \caption{Network statistics at different spatial levels}
    \begin{tabular}{|l|c|c|c|}
        \toprule
        \textbf{Level} & \textbf{Nodes} & \textbf{Edges} & \textbf{Average Weights } \\
        \midrule
        Quarters & 9 & 81 & 14,214.16 \\
        Subquarters & 31 & 914 & 1,259.68 \\
        Neighborhoods & 70 & 3,359 & 342.76 \\
        Statistical Area & 172 & 20,029 & 57.48 \\
        City Blocks & 1,205 & 45,910 & 25.08 \\
        Hexagonal Grid & 6,854 & 251,940 & 4.57 \\
        \bottomrule
    \end{tabular}
    \label{tab:graph_levels}
    \label{Network_Stats}
\end{table}

\begin{figure*}[ht]
\centering
\begin{minipage}{0.7\textwidth}
    \centering
    \includegraphics[scale=0.7]{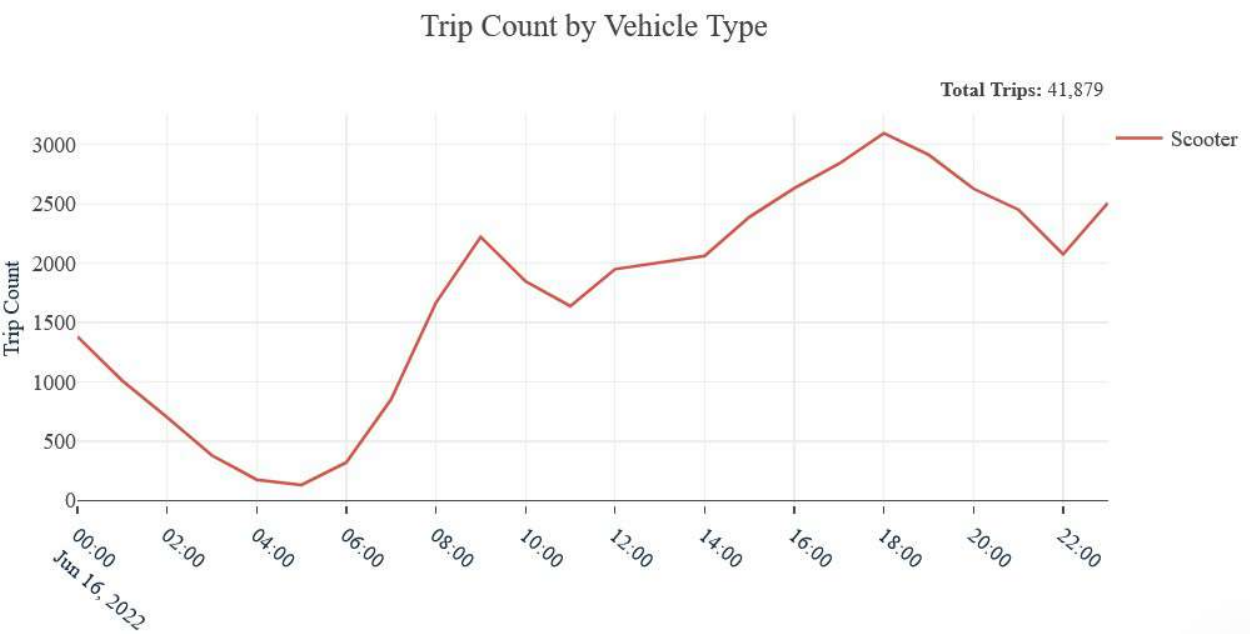}
    \label{fig40:view}
\end{minipage}%
\hfill
\begin{minipage}{0.7\textwidth}
    \centering
    \includegraphics[scale=0.7]{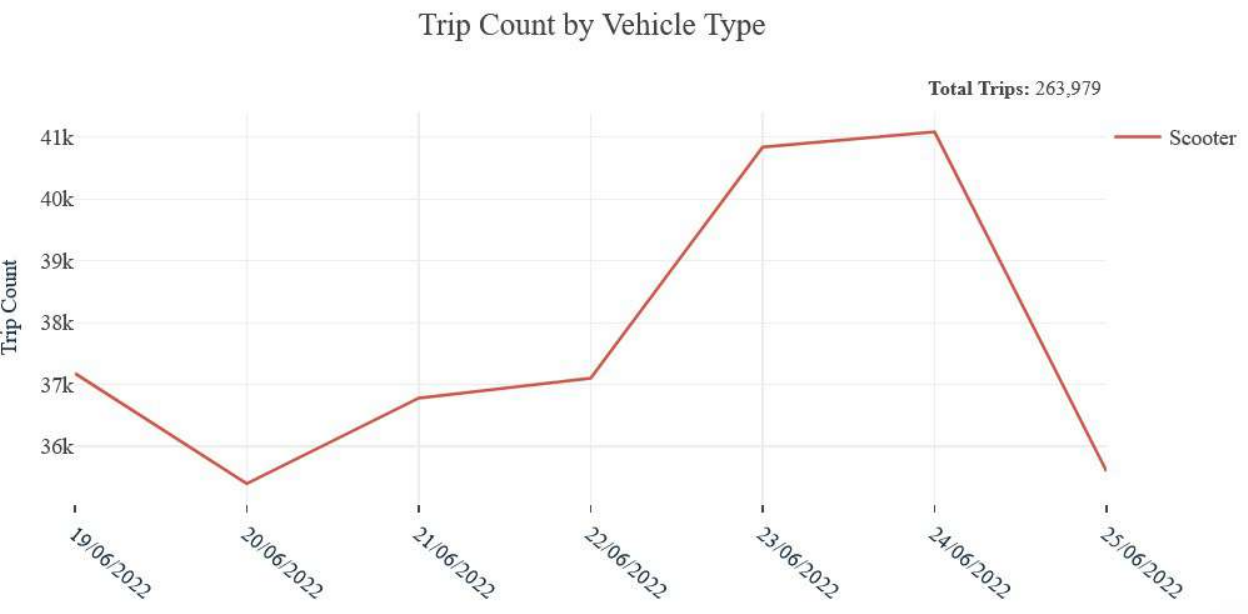}
    \label{fig41:view}
\end{minipage}

\begin{minipage}{0.7\textwidth}
    \centering
    \includegraphics[scale=0.7]{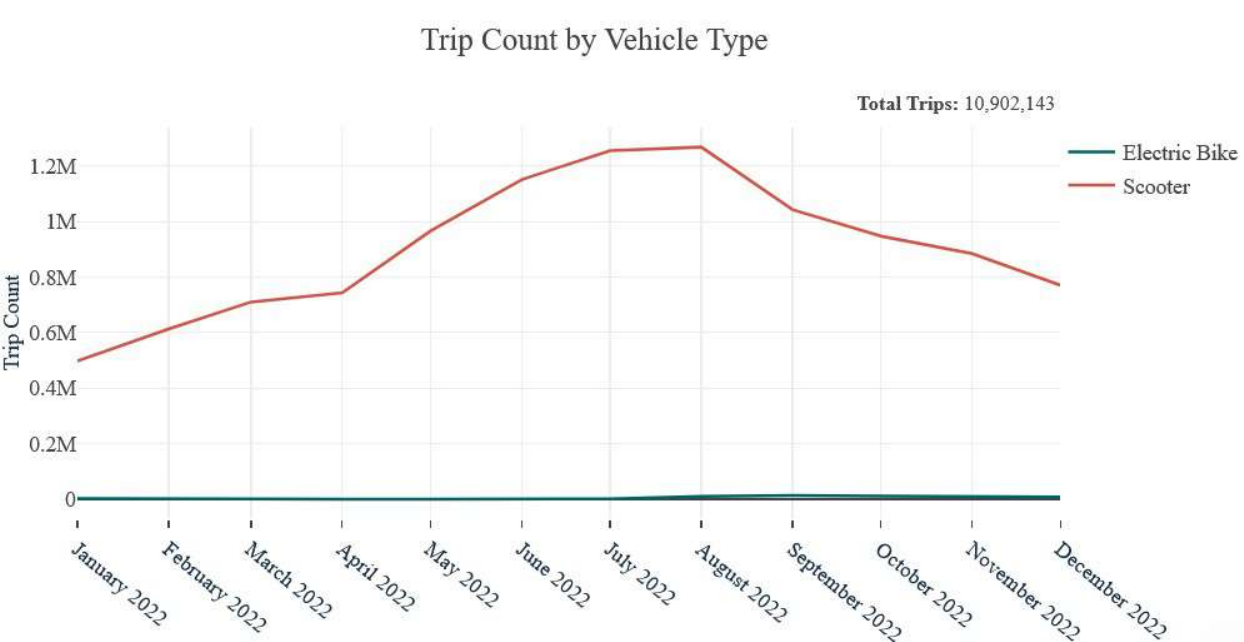}
    \label{fig42:view}
\end{minipage}
\caption{Hourly, Weekly and Yearly demand of Shared Micromobility in 2022}
\label{Demand_By_Time_Frame}
\end{figure*}

We conducted a temporal analysis of the e-scooter hourly, weekly, and yearly demand (see Figure \ref{Demand_By_Time_Frame}): We observed peak demand between 8:00 to 9:00 AM and 5:00 to 6:00 PM, aligning with commuter travel times. Minimal usage between midnight and 5:00 am, suggesting that e-scooters are for daily commute rather than late-night travel.

Then, we analyzed weekly demand in Tel Aviv on weekdays in June 2022.

We notice a higher demand on weekdays, especially Thursdays, correlates with work commute behavior. We observe a decrease in Saturdays when public transport is unavailable.

We analyze the total demand for e-scooter in Tel Aviv in 2022. We notice a significant demand increase in the summer (June–August). Winter decline, demonstrating high weather sensitivity. 
These insights indicate that e-scooter operators could dynamically adjust fleet availability in response to peak hours, weekdays, and seasons.

We conducted a spatial analysis of the e-scooter rides over three divisions- quarters, statistical areas, and blocks. 
Figure \ref{Spatial_Analysis} presents the micromobility demand by quarters in June 2022. The highest demand is in central city quarters (quarter 3 and quarter 5), characterized by commercial density and transport hubs.
It also presents the demand for micromobility by statistical areas in June 2022. The highest demand is in statistical areas 81 and 116, which have higher average income and younger population, and presents the micromobility demand by blocks in June 2022. The highest demand is in Blocks 246, 408, 1, and 831, which are near the coast.

\label{Model_Evaluations} 
We trained several machine-learning models and evaluated their performance using MAE (Mean Absolute Error), MAPE (Mean Absolute Percentage Error), and RMSE (Root Mean Squared Error) across different $T_{scale}$ and $S_{level}$. Among all machine-learning models, the XGBoost regressor \footnote{We ran the XGBoost with the following configuration: number of estimators equals to 2000, learning rate of 0.1, max depth of 5, and hist as tree method}demonstrated the best results (see Figures~\ref{fig60:view} and~\ref{fig61:view}) in almost all cases with different $T_{scale}$ and $S_{level}$, for each evaluation metric. We selected the best-performing regressor based on the importance of features and feature groups, utilizing partial feature groups and SHAP analysis.

\begin{figure*}[ht]
\centering
    \centering
    \includegraphics[scale=0.5]{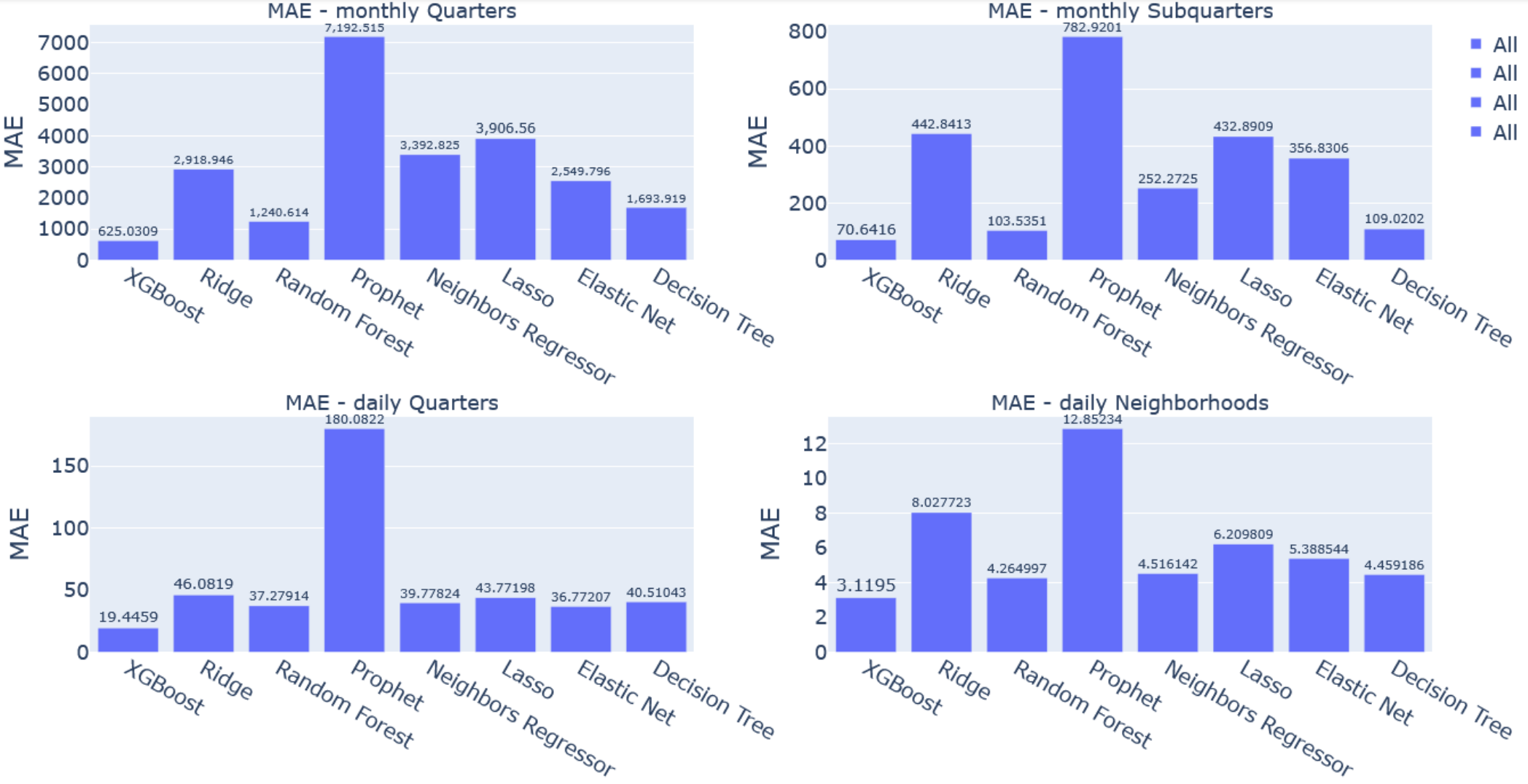}
    \caption{Comparison of MAE for various ML models, time frames and spatial divisions}
    \label{fig70:view}
    \label{Prediction_Results}
\end{figure*}

We performed a partial feature group analysis over the best-performing model (XGBoost) for each $T_{scale}$ and $S_{level}$ and evaluation metric (see Table~\ref{Table S7}). In Figure~\ref{Partial_Features_MAE} and~\ref{Partial_Features_RMSE}, we have a partial feature groups comparison for the best-performing model. 
In this analysis, network features had the highest impact on prediction accuracy, followed by temporal and spatial features.

Lastly, we performed a SHAP analysis over the best-performing model (XGBoost). Figures~\ref{SHAP_Analysis} show the 10 most prominent features for prediction, for different $T_{scale}$ and $S_{level}$. The SHAP analysis revealed that previous count, 
shortest path length, degree, centrality and strength
were the most significant factors influencing e-scooter demand. Among temporal features, the most influencing features were previous count, previous weather description, and total cases.

\section{Discussion}
\label{Discussion}
In this study, we successfully developed a general micromobility demand prediction model and tested the model in Tel Aviv. Our approach integrates machine-learning algorithms with spatial, temporal, and network features. Our research revealed that applying machine learning techniques and integrating network features enhance micromobility demand predictions, which can benefit policymakers and micromobility providers. 

According to the results, We can infer the following conclusions: 

The machine-learning algorithm generates a generic model applicable to different cities since we extracted diverse generic features (see Section~\ref{Feature_Engineering}). We improved demand predictions by extracting features from each network $G_{t}^{s}$ that efficiently represent the network topology. The demand areas are inherently imbalanced, as seen in Figure~\ref{Spatial_Analysis}. Peripheral regions exhibit lower demand, suggesting the potential to expand the accessibility of micromobility. Coastal districts (e.g., Blocks 246, 408, 1, and 831) and commercial zones exhibit high e-scooter usage due to their cycling infrastructure, proximity to the city center, and high pedestrian activity, suggesting the potential to expand micromobility adoption in those areas. Areas with higher income and younger populations (e.g., Statistical Areas 81 and 116) demonstrate higher adoption rates, aligning with prior studies on demographic influences on micromobility usage. Various feature types—spatial, temporal, and network-based—allow for a more comprehensive representation of factors influencing micromobility usage. Our framework integrates multiple geographic scales and time frames, capturing demand patterns from quarters to blocks level and from monthly to hourly timeframes. Our method requires a micromobility rides dataset, and integrating network and temporal features enhances the predictions.

Second, our model, applying the XGBoost algorithm, provided the highest performance for each $T_{scale}$ and $S_{level}$, with 27-49\% improvement in predictive accuracy compared to existing approaches and enhanced prophet predictions by 66- 91\% (see Figure~\ref{fig61:view}). The network and temporal features significantly boost model performance.

We analyze the best-performing model (XGBoost) by running partial feature groups to understand each feature group's contribution to the prediction accuracy. The partial feature groups were spatial, temporal, network, spatial and temporal, network and spatial, network and temporal, and all feature groups. 
Our results show that the combination of network and temporal features outperformed each partial feature group, including groups with spatial, temporal, and network features. We aggregated the number of facilities, surface areas, and total line lengths within each spatial division for each spatial feature. However, this approach may not fully capture the spatial context. Future research can enhance spatial feature extraction by developing more sophisticated and generalizable methods to encode spatial context.

We then evaluated each feature group separately to find the most influential feature group. Network features are the most influential, followed by temporal features and spatial attributes (see Figure ~\ref{Partial_Features_RMSE}). The network features are derived from the e-scooter rides dataset and capture key aspects of the network, such as degree and centrality, which are critical for demand prediction. The temporal features, such as weather conditions, are dynamic in time and reflect real-time fluctuations in demand. However, the spatial features we chose, such as park total area, metro line length, and buildings, might be less effective predictors of e-scooter demand since they are static in time and may not capture the spatial context. 

Spatial features demonstrated the weakest influence over predictions. It may be because the selected spatial features (see Table~\ref{Spatial_Features_Table_Tel_Aviv}) are less relevant to e-scooter movement relative to connectivity, centrality, and weather. 

The findings suggest that network features, such as centrality and connectivity, are critical in capturing micromobility patterns, particularly when combined with temporal factors, such as weather and special events, likely because temporal patterns introduce dynamic behavioral insights that complement the network features.
We achieve high prediction accuracy using network and temporal features, 103 features from the 341. However, the result is specific to the Tel Aviv network, and outcomes may differ in other cities with different infrastructures.

From the SHAP values (see Figure~\ref{SHAP_Analysis}), we notice that the characteristics of the network, mainly the centrality of the distance and the proximity, are the most influential, which means that accessibility within the micromobility network is critical in the prediction of demand. For temporal features, weather-related factors such as temperature and rainfall significantly affect usage patterns, similar to prior studies. 

However, our study has certain limitations. The model relies on historical micromobility data, which can produce biases and reduce model accuracy if the mobility patterns change over time. Although our model is general, we evaluated it in Tel Aviv. Cities with different infrastructures might show different demand patterns and require feature tuning. Generalizing the model to distinct cities can improve the robustness of the model and refine the predictions. 
We mainly focused on spatial, temporal, and network-based features. Other factors, such as user behavior and demographics, can influence micromobility demand, and integrating them can enhance model accuracy. 
Our study relies on open-source and municipal datasets, which can contain missing values, inconsistencies, or biases. The micromobility data may not always be accessible. Those factors can affect the model's predictions significantly.

\section{Conclusions}
\label{Conclusion} 

This study presents a machine-learning-based methodology to forecast micromobility demand, integrating spatial, temporal, and network-based features. Our method follows a four-step process: data collection (see Section \ref{Data_Processing}), feature extraction (see Section \ref{Feature_Engineering}), constructing a prediction model (see Section \ref{model_construction}), and model evaluation and analysis (see Section \ref{Model_Evaluation}). We used various regression-based models to predict e-scooter demand, chose the best-performing model, and performed partial feature group analysis and SHAP to find the most influential features and feature groups. 

We applied our framework to e-scooter demand prediction in Tel Aviv, Israel and evaluated the performance across different $S_{level}$ and $T_{scale}$. Using our method over different regressors, over the spatial, temporal, and network features, we notice that the XGBoost model outperforms the models for each $S_{level}$ and $T_{scale}$, reducing the second-best model’s MAE and RMSE by 27–49\%. We then chose the best-performing model (XGBoost) and evaluated it over partial feature groups. Interestingly, XGBoost models trained on temporal and network features outperformed models trained on all three feature groups. The SHAP analysis determined that the most influential network features were degree, strength, shortest path length, and weather conditions among temporal features. 


\label{Future_Work} 
There are many potential directions for future research. Micromobility providers and policymakers can derive valuable insights using our method. Applying our methods in cities such as Los Angeles, Tokyo, and Amsterdam is a promising research direction to improve generalization. 
Future studies can expand to other micromobility modes, such as e-bikes and segways, enhancing urban mobility planning. Each mode has unique operational constraints, and an integrated dataset can enable micromobility providers to coordinate between different micromobility modes and promote micromobility usage. Future studies can enrich the data to provide a more comprehensive understanding of micromobility trends. Combining datasets on special events, traffic conditions, and socio-economic factors can improve micromobility demand predictions.

We can test different machine learning techniques to enhance predictive accuracy, as mentioned in Section \ref{Temporal_ML_approaches}. Applying spatial models like Geographically Weighted Regression~\cite{brunsdon1998geographically} can address localized variations, while Graph Neural Networks~\cite{scarselli2008graph} and Spatiotemporal Graph Neural Networks~\cite{cini2023scalable} offer improved forecasting by capturing both spatial dependencies and temporal trends.

Finally, we hope to extend this research by developing a generic, flexible, open-source framework to forecast micromobility demand. This framework would enable researchers and practitioners to change and adjust the parts of our method and use different algorithms for each part, enabling further exploration and understanding of micromobility demand.

\section{Acknowledgments} 
This research was supported by Dr. Ethel Friedmann Fund.
We thank Tel Aviv Municipality and Populus AI for providing the data for this study. 
In addition, while drafting this article, we used ChatGPT, Claude and Grammarly for slight editing according to necessity.

\phantomsection
\bibliographystyle{unsrt}
\bibliography{main}

\clearpage 
\appendix  

\section*{Appendix}

\subsection{Features Tables}
\label{Appendix_Features_Table}

The mean absolute error (MAE) measures the average absolute difference between the predicted and actual values.
Mean Absolute Percentage Error (MAPE) - Measures errors as percentages to compare between scales.
Root Mean Squared Error (RMSE)\footnote {MSE: Quantifies the average squared difference between the predicted and actual values, penalizing higher deviations.}\footnote {RMSE: The square root of MSE provides an interpretable measure of error magnitude, useful for penalizing large deviations. While RMSE is sensitive to outliers, its emphasis on extensive errors makes it a valuable metric for applications where extreme mispredictions can significantly impact decision-making.}

\label{Features_Table}

Those are the spatial features we used:

\renewcommand{\thetable}{S\arabic{table}}
\onecolumn
\begin{longtable}{|p{0.2\textwidth}|p{0.5\textwidth}|p{0.2\textwidth}|}
    \caption{Table of the spatial features we used in Tel Aviv\cite{tel_aviv_open_data}:} \label{Table S1} \\
    \hline
    \textbf{Feature} & \textbf{Description} & \textbf{Dataset} \\
    \hline
    \endfirsthead
    
    \hline
     \textbf{Feature} & \textbf{Description}  & \textbf{Dataset}\\
    \hline
    \endhead
    
    \hline
    \endfoot
    
    \hline
    \endlastfoot
    \label{Spatial_Features_Table_Tel_Aviv}

        Number of Public Hosting Spaces & Total number of public areas available for hosting events or gatherings & Tel Aviv Open Data. \\ \hline
        Number of Licensed Businesses & Total number of businesses operating with a valid license or permit & Tel Aviv Open Data. \\ \hline
        Number of Nature Sites (2012) & Total number of designated nature sites as of 2012 & Tel Aviv Open Data. \\ \hline
        Area Owned by Municipality & Total land area owned by the city municipality & Tel Aviv Open Data. \\ \hline
        Number of Social Services Departments & Total number of departments providing social services within the city & Tel Aviv Open Data. \\ \hline
        Number of Public Innovation Projects & Total number of projects aimed at fostering innovation in public spaces & Tel Aviv Open Data. \\ \hline
        Area of Building Requests and Permits & Total area for which construction requests and permits have been submitted & Tel Aviv Open Data. \\ \hline
        Area of Construction Sites & Total area designated for ongoing construction projects & Tel Aviv Open Data. \\ \hline
        Number of Azor VaBitzron (Support Programs) & Total number of social support programs (e.g., Azor VaBitzron) & Tel Aviv Open Data. \\ \hline
        Number of Immigrant Absorption Centers & Total number of centers dedicated to helping new immigrants integrate into society & Tel Aviv Open Data. \\ \hline
        Number of Rights Assistance Centers & Total number of centers assisting residents in exercising their rights (e.g., legal rights, social benefits) & Tel Aviv Open Data. \\ \hline
        Number of Nursing Homes & Total number of nursing homes for the elderly & Tel Aviv Open Data. \\ \hline
        Number of Day Centers for the Elderly & Total number of day centers dedicated to elderly care & Tel Aviv Open Data. \\ \hline
        Number of Senior Citizen Clubs & Total number of social clubs for senior citizens & Tel Aviv Open Data. \\ \hline
        Number of Disability Institutions & Total number of institutions offering services and support to people with disabilities & Tel Aviv Open Data. \\ \hline
        Number of Public Toilets & Total number of public toilet facilities & Tel Aviv Open Data. \\ \hline
        Number of Drinking Fountains & Total number of public drinking fountains & Tel Aviv Open Data. \\ \hline
        Number of Dog Parks & Total number of parks designated for dog exercise and play & Tel Aviv Open Data. \\ \hline
        Number of Street Cat Feeding Stations & Total number of designated feeding stations for stray cats & Tel Aviv Open Data. \\ \hline
        Length of Ecological Corridors & Total length of designated ecological corridors aimed at preserving biodiversity & Tel Aviv Open Data. \\ \hline
        Number of Solar-Powered Rooftops & Total number of rooftops with installed solar panels & Tel Aviv Open Data. \\ \hline
        Number of Waste Collection Points & Total number of designated spots for the collection of green waste & Tel Aviv Open Data. \\ \hline
        Number of Recycling Facilities & Total number of recycling facilities available in the city & Tel Aviv Open Data. \\ \hline
        Number of Composters & Total number of composting facilities for organic waste & Tel Aviv Open Data. \\ \hline
        Number of Underground Waste Containers & Total number of underground waste collection containers & Tel Aviv Open Data. \\ \hline
        Number of Public Wi-Fi Points & Total number of public Wi-Fi access points in the city & Tel Aviv Open Data. \\ \hline
        Number of Public Notice Boards & Total number of public notice boards installed throughout the city & Tel Aviv Open Data. \\ \hline
        Number of Trees & Total number of trees in public spaces across the city & Tel Aviv Open Data. \\ \hline
        Area of Social Welfare Zones & Total area designated for welfare-related services and facilities & Tel Aviv Open Data. \\ \hline
        Area of Urban Nature Sites & Total area of sites designated for the preservation of urban nature & Tel Aviv Open Data. \\ \hline
        Number of Good Deeds Day Events & Total number of community events held on Good Deeds Day, aimed at promoting social responsibility & Tel Aviv Open Data. \\ \hline
        Metro Route M1 Area & Total area of land dedicated to the M1 metro route & Tel Aviv Open Data. \\ \hline
        Metro Route M2 Area & Total area of land dedicated to the M2 metro route & Tel Aviv Open Data. \\ \hline
        Metro Route M3 Area & Total area of land dedicated to the M3 metro route & Tel Aviv Open Data. \\ \hline
        Number of Purple Line Stations & Total number of stations along the Purple Line of the light rail system & Tel Aviv Open Data. \\ \hline
        Number of Green Line Stations & Total number of stations along the Green Line of the light rail system & Tel Aviv Open Data. \\ \hline
        Number of Red Line Stations & Total number of stations along the Red Line of the light rail system & Tel Aviv Open Data. \\ \hline
        Purple Line Track Area & Total land area allocated for the Purple Line track & Tel Aviv Open Data. \\ \hline
        Green Line Track Area & Total land area allocated for the Green Line track & Tel Aviv Open Data. \\ \hline
        Red Line Track Area & Total land area allocated for the Red Line track & Tel Aviv Open Data. \\ \hline
        Light Rail Engineering Facilities Area & Total area dedicated to the engineering facilities for the light rail system & Tel Aviv Open Data. \\ \hline
        Light Rail Staging Areas & Total land area used for light rail construction staging & Tel Aviv Open Data. \\ \hline
        Number of Loading/Unloading Signs & Total number of signs designated for loading and unloading zones & Tel Aviv Open Data. \\ \hline
        Number of Waze Reports & Total number of traffic or incident reports generated through Waze & Tel Aviv Open Data. \\ \hline
        Length of Waze Traffic Congestion & Total length of traffic congestion areas reported via Waze & Tel Aviv Open Data. \\ \hline
        Number of Night Parking Lots for Residents & Total number of parking lots dedicated to residents for overnight parking & Tel Aviv Open Data. \\ \hline
        Number of Ahuzat HaHof Parking Lots & Total number of parking lots managed by Ahuzat HaHof (the municipal parking authority) & Tel Aviv Open Data. \\ \hline
        Area for Limited Access Zones for Shared Vehicles & Total area where access is restricted or regulated for shared vehicle services & Tel Aviv Open Data. \\ \hline
        Number of Parking Spots for Micro-Mobility Vehicles & Total number of parking spaces designated for small electric vehicles (e.g., scooters) & Tel Aviv Open Data. \\ \hline
        Length of Pedestrian-Friendly Streets & Total length of streets that prioritize pedestrian movement and walkability & Tel Aviv Open Data. \\ \hline
        Number of AutoTel Parking Spots & Total number of parking spots reserved for AutoTel (a car-sharing service) & Tel Aviv Open Data. \\ \hline
        Number of Tel-O-Fun Bike Stations & Total number of docking stations for the Tel-O-Fun bike-sharing program & Tel Aviv Open Data. \\ \hline
        Number of Hop-On Charging Stations & Total number of charging stations for the Hop-On electric mobility service & Tel Aviv Open Data. \\ \hline
        Number of Public Transit Lane Cameras & Total number of cameras monitoring lanes designated for public transit & Tel Aviv Open Data. \\ \hline
        Number of NETZ Cameras & Total number of NETZ cameras used for traffic enforcement & Tel Aviv Open Data. \\ \hline
        Number of Public Transit Terminals & Total number of public transit terminals & Tel Aviv Open Data. \\ \hline
        Number of Weekend Public Transit Stations & Total number of stations where weekend public transit is available & Tel Aviv Open Data. \\ \hline
        Length of Weekend Public Transit Lines & Total length of routes served by weekend public transit & Tel Aviv Open Data. \\ \hline
        Number of Ministry of Transportation Bus Stops & Total number of bus stops managed by the Ministry of Transportation & Tel Aviv Open Data. \\ \hline
        Number of Traffic Lights at Intersections & Total number of intersections with traffic lights & Tel Aviv Open Data. \\ \hline
        Area of Night Construction - Buildings & Total area where night construction activities related to buildings are occurring & Tel Aviv Open Data. \\ \hline
        Number of Public Parking Lots & Total number of public parking lots & Tel Aviv Open Data. \\ \hline
        Number of Private Parking Lots & Total number of privately owned parking lots & Tel Aviv Open Data. \\ \hline
        Area of Bubble Dan Service Zones & Total area where the Bubble Dan on-demand public transportation service operates & Tel Aviv Open Data. \\ \hline
        Number of Loading/Unloading Pilots & Total number of pilot programs for designated loading and unloading zones & Tel Aviv Open Data. \\ \hline
        Number of Motorcycle Parking Spaces & Total number of parking spaces allocated for motorcycles & Tel Aviv Open Data. \\ \hline
        Number of Traffic Signs & Total number of traffic direction signs in the city & Tel Aviv Open Data. \\ \hline
        Number of Fuel Stations & Total number of fuel stations within the city & Tel Aviv Open Data. \\ \hline
        Number of Intersections & Total number of road intersections & Tel Aviv Open Data. \\ \hline
        Number of Bicycle Maintenance Stations & Total number of bicycle maintenance stations throughout the city & Tel Aviv Open Data. \\ \hline
        Length of Bicycle Lanes & Total length of designated bicycle lanes & Tel Aviv Open Data. \\ \hline
        Length of 2025 City Bicycle Strategy & Total length of bicycle lanes planned under the city's 2025 bicycle strategy & Tel Aviv Open Data. \\ \hline
        Length of Bicycle Strategy Execution Status & Total length of bicycle lanes that have been completed as part of the 2025 strategy & Tel Aviv Open Data. \\ \hline
        Length of the Ophnidan (Bicycle Highway) & Total length of the Ophnidan, a high-speed bicycle route designed for urban commuting & Tel Aviv Open Data. \\ \hline
        Length of Bridges & Total length of bridges in the city & Tel Aviv Open Data. \\ \hline
        Length of Public Transit Lanes & Total length of dedicated public transit lanes & Tel Aviv Open Data. \\ \hline
        Area of Vehicle Towing Zones & Total area where vehicles are towed due to illegal parking or other violations & Tel Aviv Open Data. \\ \hline
        Area of Parking Zones & Total area designated for public parking & Tel Aviv Open Data. \\ \hline
        Number of Defibrillators & Total number of publicly available defibrillators & Tel Aviv Open Data. \\ \hline
        Number of Medical Institutions & Total number of medical institutions, including hospitals and clinics & Tel Aviv Open Data. \\ \hline
        Number of Health Clinics & Total number of general health clinics in the city & Tel Aviv Open Data. \\ \hline
        Number of Pharmacies & Total number of pharmacies & Tel Aviv Open Data. \\ \hline
        Number of Tipat Halav Clinics & Total number of Tipat Halav (maternal and child health) clinics & Tel Aviv Open Data. \\ \hline
        Number of Mental Health Services & Total number of mental health facilities and services & Tel Aviv Open Data. \\ \hline
        Number of Magen David Adom Stations & Total number of emergency stations operated by Magen David Adom & Tel Aviv Open Data. \\ \hline
        Number of Educational Psychological Services & Total number of stations offering educational psychological services & Tel Aviv Open Data. \\ \hline
        Area of Traffic Noise Levels (4 meters) & Total area exposed to traffic noise at a height of 4 meters above ground level & Tel Aviv Open Data. \\ \hline
        Area of Traffic Noise Levels (45 meters) & Total area exposed to traffic noise at a height of 45 meters above ground level & Tel Aviv Open Data. \\ \hline
        Number of Cell Towers Under Construction & Total number of cellular towers that are under construction & Tel Aviv Open Data. \\ \hline
        Number of Operational Cell Towers & Total number of fully operational cellular towers & Tel Aviv Open Data. \\ \hline
        Number of Tashpe Schools (Elementary) & Total number of elementary schools categorized as Tashpe (state-religious) & Tel Aviv Open Data. \\ \hline
        Number of Tashpad Schools (Elementary) & Total number of elementary schools categorized as Tashpad (state-secular) & Tel Aviv Open Data. \\ \hline
        Number of Tashpe Kindergartens & Total number of kindergartens categorized as Tashpe (state-religious) & Tel Aviv Open Data. \\ \hline
        Number of Tashpad Kindergartens & Total number of kindergartens categorized as Tashpad (state-secular) & Tel Aviv Open Data. \\ \hline
        Number of Youth Clubs & Total number of youth clubs & Tel Aviv Open Data. \\ \hline
        Number of Recognized Daycares & Total number of officially recognized early childhood daycare centers & Tel Aviv Open Data. \\ \hline
        Number of School District Labels (Tashpe) & Total number of school district labels for Tashpe schools & Tel Aviv Open Data. \\ \hline
        Number of School District Labels (Tashpad) & Total number of school district labels for Tashpad schools & Tel Aviv Open Data. \\ \hline
        Number of General School District Labels & Total number of school district labels for general schools & Tel Aviv Open Data. \\ \hline
        Number of State School District Labels & Total number of school district labels for state schools & Tel Aviv Open Data. \\ \hline
        Number of Religious School District Labels & Total number of school district labels for religious schools & Tel Aviv Open Data. \\ \hline
        Number of Arab School District Labels & Total number of school district labels for Arab schools & Tel Aviv Open Data. \\ \hline
        Number of Synagogues & Total number of synagogues & Tel Aviv Open Data. \\ \hline
        Number of Mikvahs (Ritual Baths) & Total number of mikvahs (ritual baths) & Tel Aviv Open Data. \\ \hline
        Number of Cemeteries & Total number of cemeteries & Tel Aviv Open Data. \\ \hline
        Number of Other Religious Institutions & Total number of religious institutions other than synagogues or cemeteries & Tel Aviv Open Data. \\ \hline
        Area of Eruv Lines & Total length or area covered by Eruv lines (boundaries for religious purposes) & Tel Aviv Open Data. \\ \hline
        Number of Tourist Offices & Total number of tourist information centers or offices & Tel Aviv Open Data. \\ \hline
        Number of ATMs & Total number of ATMs (automated teller machines) & Tel Aviv Open Data. \\ \hline
        Number of Bank Branches & Total number of bank branches & Tel Aviv Open Data. \\ \hline
        Beach Area & Total area designated as beach property & Tel Aviv Open Data. \\ \hline
        Green Space Area and Public Parks & Total area of green spaces and public parks & Tel Aviv Open Data. \\ \hline
        Number of Embassies & Total number of embassies & Tel Aviv Open Data. \\ \hline
        Number of Hotels & Total number of hotels & Tel Aviv Open Data. \\ \hline
        Number of Playgrounds & Total number of playgrounds for children & Tel Aviv Open Data. \\ \hline
        Number of Community Gardens & Total number of community gardens & Tel Aviv Open Data. \\ \hline
        Number of Water Fountains & Total number of water fountains & Tel Aviv Open Data. \\ \hline
        Number of Table Tennis Facilities in Parks & Total number of table tennis facilities located in public parks & Tel Aviv Open Data. \\ \hline
        Number of Playrooms & Total number of indoor playrooms for children & Tel Aviv Open Data. \\ \hline
        Number of Fountains & Total number of public fountains & Tel Aviv Open Data. \\ \hline
        Number of Community Institutions & Total number of community institutions & Tel Aviv Open Data. \\ \hline
        Number of Post Offices & Total number of post offices & Tel Aviv Open Data. \\ \hline
        Number of Stadiums and Sports Halls & Total number of stadiums and sports halls & Tel Aviv Open Data. \\ \hline
        Number of Gyms & Total number of gyms or fitness centers & Tel Aviv Open Data. \\ \hline
        Number of Sports Halls & Total number of indoor sports halls & Tel Aviv Open Data. \\ \hline
        Number of Swimming Pools & Total number of public swimming pools & Tel Aviv Open Data. \\ \hline
        Number of Sports Fields & Total number of outdoor sports fields & Tel Aviv Open Data. \\ \hline
        Number of Fitness Facilities in Parks & Total number of fitness facilities located in public parks & Tel Aviv Open Data. \\ \hline
        Number of Beach and Ocean Sports Facilities & Total number of sports facilities located on the beach or in the ocean & Tel Aviv Open Data. \\ \hline
        Number of Donation Benches & Total number of benches designated for charitable purposes (e.g., benches with plaques recognizing donations) & Tel Aviv Open Data. \\ \hline
        Number of Special Trees & Total number of trees designated as special or protected & Tel Aviv Open Data. \\ \hline
        Number of This-Is-The-Place Sites & Total number of sites designated under ""This-Is-The-Place"" for historical or cultural reasons & Tel Aviv Open Data. \\ \hline
        Number of Youth Movements & Total number of youth movements or organizations & Tel Aviv Open Data. \\ \hline
        Number of Young Adult and Entrepreneurship Centers & Total number of centers focused on young adults and entrepreneurship & Tel Aviv Open Data. \\ \hline
        Number of Dance Centers & Total number of centers focused on dance or performing arts & Tel Aviv Open Data. \\ \hline
        Number of Libraries & Total number of public libraries & Tel Aviv Open Data. \\ \hline
        Number of Outdoor Libraries & Total number of outdoor libraries & Tel Aviv Open Data. \\ \hline
        Number of Memorials & Total number of memorials or monuments & Tel Aviv Open Data. \\ \hline
        Number of Museums & Total number of museums & Tel Aviv Open Data. \\ \hline
        Number of Art Galleries & Total number of art galleries & Tel Aviv Open Data. \\ \hline
        Number of Theaters & Total number of theaters or performance spaces & Tel Aviv Open Data. \\ \hline
        Number of Cinemas & Total number of movie theaters & Tel Aviv Open Data. \\ \hline
        Number of Music Centers & Total number of centers focused on music education or performance & Tel Aviv Open Data. \\ \hline
        Number of Auditoriums & Total number of auditoriums & Tel Aviv Open Data. \\ \hline
        Area of Sustainable Neighborhoods & Total area of neighborhoods classified as sustainable or eco-friendly & Tel Aviv Open Data. \\ \hline
        Number of Green Schools & Total number of schools certified as green or eco-friendly & Tel Aviv Open Data. \\ \hline
        Length of Tel Aviv Marathon (42.2 km) & Total length of the Tel Aviv marathon route (42.2 kilometers) & Tel Aviv Open Data. \\ \hline
\end{longtable}

\renewcommand{\thetable}{S\arabic{table}}
\onecolumn
\begin{longtable}{|p{0.2\textwidth}|p{0.5\textwidth}|p{0.2\textwidth}|}
    \caption{Table of the spatial features we used in the methods:} \label{tab1:long} \\
    \hline
    \textbf{Feature} & \textbf{Description}
    & \textbf{Dataset}
    \\
    \hline
    \endfirsthead
    
    \hline
     \textbf{Feature} & \textbf{Description}
     & \textbf{Dataset}
     \\
    \hline
    \endhead
    
    \hline
    \endfoot
    
    \hline
    \endlastfoot
    \label{Spatial_Features_Table}

        Number of Public Hosting Spaces & Total number of public areas available for hosting events or gatherings
        & Open GIS layers Datasets.\\ \hline
        Number of Licensed Businesses & Total number of businesses operating with a valid license or permit
        & Open GIS layers Datasets.\\ \hline
        Number of Nature Sites & Total number of designated nature sites 
        & Open GIS layers Datasets.\\ \hline
        Area Owned by Municipality & Total land area owned by the city municipality  & Open GIS layers Datasets.
        \\ \hline
        Number of Social Services Departments & Total number of departments providing social services within the city 
        & Open GIS layers Datasets.\\ \hline
        Number of Public Innovation Projects & Total number of projects aimed at fostering innovation in public spaces 
        & Open GIS layers Datasets.\\ \hline
        Area of Building Requests and Permits & Total area for which construction requests and permits have been submitted 
        & Open GIS layers Datasets.\\ \hline
        Area of Construction Sites & Total area designated for ongoing construction projects
        & Open GIS layers Datasets.\\ \hline
        Number of Immigrant Absorption Centers & Total number of centers dedicated to helping new immigrants integrate into society 
        & Open GIS layers Datasets.\\ \hline
        Number of Rights Assistance Centers & Total number of centers assisting residents in exercising their rights (e.g., legal rights, social benefits)
        & Open GIS layers Datasets.
        \\ \hline
        Number of Nursing Homes & Total number of nursing homes for the elderly  
        & Open GIS layers Datasets.\\ \hline
        Number of Day Centers for the Elderly & Total number of day centers dedicated to elderly care
        & Open GIS layers Datasets.\\ \hline
        Number of Public Toilets & Total number of public toilet facilities
        & Open GIS layers Datasets.
        \\ \hline
        Number of Drinking Fountains & Total number of public drinking fountains
        & Open GIS layers Datasets.\\ \hline
        Number of Dog Parks & Total number of parks designated for dog exercise and play 
        & Open GIS layers Datasets.
        \\ \hline
        Number of Street Cat Feeding Stations & Total number of designated feeding stations for stray cats
        & Open GIS layers Datasets.
        \\ \hline
        Length of Ecological Corridors & Total length of designated ecological corridors aimed at preserving biodiversity
        & Open GIS layers Datasets.
        \\ \hline
        Number of Solar-Powered Rooftops & Total number of rooftops with installed solar panels
        & Open GIS layers Datasets.
        \\ \hline
        Number of Waste Collection Points & Total number of designated spots for the collection of green waste & Open GIS layers Datasets.
        \\ \hline
        Number of Recycling Facilities & Total number of recycling facilities available in the city 
        & Open GIS layers Datasets.
        \\ \hline
        Number of Composters & Total number of composting facilities for organic waste
        & Open GIS layers Datasets.
        \\ \hline
        Number of Underground Waste Containers & Total number of underground waste collection containers
        & Open GIS layers Datasets.
        \\ \hline
        Number of Public Wi-Fi Points & Total number of public Wi-Fi access points in the city 
        & Open GIS layers Datasets.
        \\ \hline
        Number of Public Notice Boards & Total number of public notice boards installed throughout the city 
        & Open GIS layers Datasets.
        \\ \hline
        Number of Trees & Total number of trees in public spaces across the city & Open GIS layers Datasets.
        \\ \hline
        Area of Social Welfare Zones & Total area designated for welfare-related services and facilities & Open GIS layers Datasets.
        \\ \hline
        Area of Urban Nature Sites & Total area of sites designated for the preservation of urban nature 
        & Open GIS layers Datasets.
        \\ \hline
        Metro Route Area & Total area of land dedicated to the metro 
        & Open GIS layers Datasets.
        \\ \hline
        Light Rail Engineering Facilities Area & Total area dedicated to the engineering facilities for the light rail system & Open GIS layers Datasets.
        \\ \hline
        Light Rail Staging Areas & Total land area used for light rail construction staging
        & Open GIS layers Datasets.
        \\ \hline
        Number of Loading/Unloading Signs & Total number of signs designated for loading and unloading zones & Open GIS layers Datasets.
        \\ \hline
        Area for Limited Access Zones for Shared Vehicles & Total area where access is restricted or regulated for shared vehicle services 
        & Open GIS layers Datasets.
        \\ \hline
        Number of Parking Spots for Micro-Mobility Vehicles & Total number of parking spaces designated for small electric vehicles (e.g., scooters) 
        & Open GIS layers Datasets.
        \\ \hline
        Length of Pedestrian-Friendly Streets & Total length of streets that prioritize pedestrian movement and walkability & Open GIS layers Datasets.
        \\ \hline
        Number of Public Transit Terminals & Total number of public transit terminals & Open GIS layers Datasets.
        \\ \hline
        Length of Weekend Public Transit Lines & Total length of routes served by weekend public transit & Open GIS layers Datasets.
        \\ \hline
        Number of Ministry of Transportation Bus Stops & Total number of bus stops managed by the Ministry of Transportation  
        & Open GIS layers Datasets.
        \\ \hline
        Number of Traffic Lights at Intersections & Total number of intersections with traffic lights 
        & Open GIS layers Datasets.
        \\ \hline
        Area of Night Construction - Buildings & Total area where night construction activities related to buildings are occurring 
        & Open GIS layers Datasets.
        \\ \hline
        Number of Public Parking Lots & Total number of public parking lots  & Open GIS layers Datasets.
        \\ \hline
        Number of Private Parking Lots & Total number of privately owned parking lots & Open GIS layers Datasets.
        \\ \hline
        Number of Motorcycle Parking Spaces & Total number of parking spaces allocated for motorcycles
        & Open GIS layers Datasets.
        \\ \hline
        Number of Traffic Signs & Total number of traffic direction signs in the city 
        & Open GIS layers Datasets.
        \\ \hline
        Number of Fuel Stations & Total number of fuel stations within the city 
        & Open GIS layers Datasets.
        \\ \hline
        Number of Intersections & Total number of road intersections  
        & Open GIS layers Datasets.
        \\ \hline
        Number of Bicycle Maintenance Stations & Total number of bicycle maintenance stations throughout the city 
        & Open GIS layers Datasets.
        \\ \hline
        Length of Bicycle Lanes & Total length of designated bicycle lanes 
        & Open GIS layers Datasets.
        \\ \hline
        Length of Bridges & Total length of bridges in the city  
        & Open GIS layers Datasets.
        \\ \hline
        Length of Public Transit Lanes & Total length of dedicated public transit lanes  
        & Open GIS layers Datasets.
        \\ \hline
        Area of Vehicle Towing Zones & Total area where vehicles are towed due to illegal parking or other violations
        & Open GIS layers Datasets.
        \\ \hline
        Area of Parking Zones & Total area designated for public parking 
        & Open GIS layers Datasets.
        \\ \hline
        Number of Medical Institutions & Total number of medical institutions, including hospitals and clinics 
        & Open GIS layers Datasets.
        \\ \hline
        Number of Health Clinics & Total number of general health clinics in the city  
        & Open GIS layers Datasets.
        \\ \hline
        Number of Pharmacies & Total number of pharmacies
        & Open GIS layers Datasets.
        \\ \hline
        Number of Mental Health Services & Total number of mental health facilities and services  & Open GIS layers Datasets.
        \\ \hline
        Emergency stations Stations & Total number of emergency stations & Open GIS layers Datasets.
        \\ \hline
        Number of Educational Psychological Services & Total number of stations offering educational psychological services & Open GIS layers Datasets.
        \\ \hline
        Number of Youth Clubs & Total number of youth clubs & Open GIS layers Datasets.
        \\ \hline
        Number of Recognized Daycares & Total number of officially recognized early childhood daycare centers & Open GIS layers Datasets.
        \\ \hline
        Number of General School District Labels & Total number of school district labels for general schools & Open GIS layers Datasets.
        \\ \hline
        Number of State School District Labels & Total number of school district labels for state schools  
        & Open GIS layers Datasets.
        \\ \hline
        Number of ATMs & Total number of ATMs (automated teller machines) & Open GIS layers Datasets.
        \\ \hline
        Number of Bank Branches & Total number of bank branches  & Open GIS layers Datasets.
        \\ \hline
        Beach Area & Total area designated as beach property  & Open GIS layers Datasets.
        \\ \hline
        Number of Embassies & Total number of embassies 
        
        & Open GIS layers Datasets.
        \\ \hline
        Number of Hotels & Total number of hotels & Open GIS layers Datasets.
        \\ \hline
        Number of Playgrounds & Total number of playgrounds for children & Open GIS layers Datasets. 
        \\ \hline
        Number of Community Gardens & Total number of community gardens  & Open GIS layers Datasets.\\ \hline
        Number of Water Fountains & Total number of water fountains& Open GIS layers Datasets.  \\ \hline
        Number of Table Tennis Facilities in Parks & Total number of table tennis facilities located in public parks  & Open GIS layers Datasets.\\ \hline
        Number of Playrooms & Total number of indoor playrooms for children & Open GIS layers Datasets.\\ \hline
        Number of Fountains & Total number of public fountains& Open GIS layers Datasets.  \\ \hline
        Number of Community Institutions & Total number of community institutions & Open GIS layers Datasets. \\ \hline
        Number of Post Offices & Total number of post offices  & Open GIS layers Datasets. \\ \hline
        Number of Stadiums and Sports Halls & Total number of stadiums and sports halls & Open GIS layers Datasets. \\ \hline
        Number of Gyms & Total number of gyms or fitness centers & Open GIS layers Datasets. \\ \hline
        Number of Sports Halls & Total number of indoor sports halls & Open GIS layers Datasets. \\ \hline
        Number of Swimming Pools & Total number of public swimming pools & Open GIS layers Datasets. \\ \hline
        Number of Sports Fields & Total number of outdoor sports fields & Open GIS layers Datasets. \\ \hline
        Number of Fitness Facilities in Parks & Total number of fitness facilities located in public parks & Open GIS layers Datasets. \\ \hline
        Number of Beach and Ocean Sports Facilities & Total number of sports facilities located on the beach or in the ocean  & Open GIS layers Datasets. \\ \hline
        Number of Youth Movements & Total number of youth movements or organizations  & Open GIS layers Datasets.\\ \hline
        Number of Young Adult and Entrepreneurship Centers & Total number of centers focused on young adults and entrepreneurship & Open GIS layers Datasets. \\ \hline
        Number of Dance Centers & Total number of centers focused on dance or performing arts & Open GIS layers Datasets. \\ \hline
        Number of  Libraries & Total number of  libraries 
        & Open GIS layers Datasets.
        \\ \hline
        Number of Memorials & Total number of memorials or monuments & Open GIS layers Datasets.
        \\ \hline
        Number of Museums & Total number of museums& Open GIS layers Datasets. \\ \hline
        Number of Art Galleries & Total number of art galleries  
        & Open GIS layers Datasets.
        \\ \hline
        Number of Theaters & Total number of theaters or performance spaces & Open GIS layers Datasets.
        \\ \hline
        Number of Cinemas & Total number of movie theaters & Open GIS layers Datasets. \\ \hline
        Number of Music Centers & Total number of centers focused on music education or performance
        & Open GIS layers Datasets.
        \\ \hline
        Number of Auditoriums & Total number of auditoriums
        & Open GIS layers Datasets.\\ \hline
        Area of  Neighborhoods & Total area of neighborhoods  & Open GIS layers Datasets.
        \\ \hline
        Number of  Schools & Total number of schools  
        & Open GIS layers Datasets.
        \\ \hline
\end{longtable}

Those are the temporal features we used:

\renewcommand{\thetable}{S\arabic{table}}
\onecolumn
\begin{longtable}{|p{0.2\textwidth}|p{0.5\textwidth}|p{0.2\textwidth}|}
    \caption{Table of the temporal features we used \cite{owid-coronavirus, owid-covid-mobility-trends, ourworldindata_crude_oil_prices}:} \label{tab2:long} \\
    \hline
    \textbf{Feature} & \textbf{Description} & \textbf{Dataset} \\
    \hline
    \endfirsthead
    
    \hline
     \textbf{Feature} & \textbf{Description}  & \textbf{Dataset}\\
    \hline
    \endhead
    
    \hline
    \endfoot
    
    \hline
    \endlastfoot
    \label{Temporal_Features_Table_Tel_Aviv}

        Total cases & The cumulative number of COVID-19 cases in a given geographic region & COVID-19. \\ \hline
        New cases & The number of new COVID-19 cases reported in the last time period & COVID-19. \\ \hline
        New cases smoothed & A smoothed representation of new COVID-19 cases to account for reporting variations & COVID-19. \\ \hline
        Total deaths & The cumulative number of deaths attributed to COVID-19 & COVID-19. \\ \hline
        New deaths & The number of new COVID-19-related deaths reported in the last time period & COVID-19. \\ \hline
        New deaths smoothed & A smoothed representation of new deaths to account for reporting variations & COVID-19. \\ \hline
        Total cases per million & The cumulative number of COVID-19 cases per million people in the population & COVID-19. \\ \hline
        New cases per million & The number of new COVID-19 cases per million people in the population & COVID-19. \\ \hline
        New cases smoothed per million & A smoothed representation of new cases per million people & COVID-19. \\ \hline
        Total deaths per million & The cumulative number of deaths per million people in the population & COVID-19. \\ \hline
        New deaths per million & The number of new COVID-19 deaths per million people & COVID-19. \\ \hline
        New deaths smoothed per million & A smoothed representation of new deaths per million people & COVID-19. \\ \hline
        Reproduction rate & The effective reproduction number (R), representing how many people one infected person will infect & COVID-19. \\ \hline
        ICU patients & The number of patients currently in intensive care units due to COVID-19 & COVID-19. \\ \hline
        ICU patients per million & The number of ICU patients per million people in the population & COVID-19. \\ \hline
        Hospital patients & The number of COVID-19 patients currently hospitalized & COVID-19. \\ \hline
        Hospital patients per million & The number of hospitalized COVID-19 patients per million people & COVID-19. \\ \hline
        Weekly ICU admissions & The number of new ICU admissions over the past week & COVID-19. \\ \hline
        Weekly ICU admissions per million & The number of new ICU admissions per million people over the past week & COVID-19. \\ \hline
        Weekly hospital admissions & The number of new hospital admissions over the past week & COVID-19. \\ \hline
        Weekly hospital admissions per million & The number of new hospital admissions per million people over the past week & COVID-19. \\ \hline
        Total tests & The cumulative number of COVID-19 tests performed & COVID-19. \\ \hline
        New tests & The number of new COVID-19 tests conducted during the last time period & COVID-19. \\ \hline
        Total tests per thousand & The cumulative number of COVID-19 tests per thousand people in the population & COVID-19. \\ \hline
        New tests per thousand & The number of new COVID-19 tests per thousand people in the population & COVID-19. \\ \hline
        New tests smoothed & A smoothed representation of new tests to account for reporting variations & COVID-19. \\ \hline
        New tests smoothed per thousand & A smoothed representation of new tests per thousand people & COVID-19. \\ \hline
        Positive rate & The percentage of COVID-19 tests that return a positive result & COVID-19. \\ \hline
        Tests per case & The number of tests conducted for each reported COVID-19 case & COVID-19. \\ \hline
        Total vaccinations & The cumulative number of COVID-19 vaccinations administered & COVID-19. \\ \hline
        People vaccinated & The number of people who have received at least one dose of the COVID-19 vaccine & COVID-19. \\ \hline
        People fully vaccinated & The number of people who have received all doses required to be considered fully vaccinated & COVID-19. \\ \hline
        Total boosters & The total number of COVID-19 booster doses administered & COVID-19. \\ \hline
        New vaccinations & The number of new COVID-19 vaccinations administered in the last time period & COVID-19. \\ \hline
        New vaccinations smoothed & A smoothed representation of new vaccinations to account for reporting variations & COVID-19. \\ \hline
        Total vaccinations per hundred & The cumulative number of COVID-19 vaccinations administered per hundred people in the population & COVID-19. \\ \hline
        People vaccinated per hundred & The number of people who have received at least one dose of the vaccine per hundred people in the population & COVID-19. \\ \hline
        People fully vaccinated per hundred & The number of fully vaccinated people per hundred people in the population & COVID-19. \\ \hline
        Total boosters per hundred & The number of booster doses administered per hundred people in the population & COVID-19. \\ \hline
        New vaccinations smoothed per million & A smoothed representation of new vaccinations per million people & COVID-19. \\ \hline
        New people vaccinated smoothed & A smoothed representation of new people vaccinated & COVID-19. \\ \hline
        New people vaccinated smoothed per hundred & A smoothed representation of new people vaccinated per hundred people in the population & COVID-19. \\ \hline
        Stringency index & A composite measure of the severity of government-imposed COVID-19 measures, such as lockdowns and travel restrictions & COVID-19. \\ \hline
        Handwashing facilities & The availability of handwashing facilities in a region, important for COVID-19 mitigation & COVID-19. \\ \hline
        Excess mortality cumulative absolute & The total number of deaths above what would normally be expected during the pandemic & COVID-19. \\ \hline
        Excess mortality cumulative & The percentage increase in mortality above what would be expected during the pandemic & COVID-19. \\ \hline
        Excess mortality & The number of deaths above the expected number during a specific period & COVID-19. \\ \hline
        Excess mortality cumulative per million & The number of excess deaths per million people & COVID-19. \\ \hline
\end{longtable}

Those are the network features we used:

\renewcommand{\thetable}{S\arabic{table}}
\onecolumn
\begin{longtable}{|p{0.2\textwidth}|p{0.5\textwidth}|p{0.2\textwidth}| }
    \caption{Table of the network features we used\cite{populusAIRidesData, hagberg2008exploring}:} \label{tab3:long} \\
    \hline
    \textbf{Feature} & \textbf{Description} & \textbf{Dataset} \\
    \hline
    \endfirsthead
    
    \hline
     \textbf{Feature} & \textbf{Description}  & \textbf{Dataset}\\
    \hline
    \endhead
    
    \hline
    \endfoot
    
    \hline
    \endlastfoot
    \label{Network_Features_Table}

        Degree centrality & The degree centrality for a node v is the fraction of nodes it is connected to. & Network data. \\ \hline
        In degree centrality & The in-degree centrality for a node v is the fraction of nodes its incoming edges are connected to. & Network data. \\ \hline
        Out degree centrality & The out-degree centrality for a node v is the fraction of nodes its outgoing edges are connected to. & Network data. \\ \hline
        Betweenness centrality & Betweenness centrality of a node is the sum of the fraction of all-pairs shortest paths that pass through & Network data. \\\hline
        In degree & The node $in_{deg}$ is the number of edges pointing to the node. The weighted node degree is the sum of the edge weights for edges incident to that node. & Network data.\\ \hline
        Out degree & The node $out_{deg}$ is the number of edges pointing out of the node. The weighted node degree is the sum of the edge weights for edges incident to that node. & Network data. \\ \hline
        Edge betweenness & Betweenness centrality of an edge is the sum of the fraction of all-pairs shortest paths that pass through the edge & Network data. \\ \hline
        Shortest paths length & Compute shortest path lengths in the graph. & Network data. \\ \hline
        Number of nodes & Returns the number of nodes in the graph. & Network data. \\ \hline
        Number of edges & Returns the number of edges between two nodes. & Network data. \\ \hline
        Average Degree & Compute the average degree connectivity of graph. The average degree connectivity is the average nearest neighbor degree of nodes with degree k. For weighted graphs, an analogous measure can be computed using the weighted average neighbors degree defined in [1], for a node i, as where $s_i$ is the weighted degree of node i, $w_{ij}$ is the weight of the edge that links i and j, and $N(i)$ are the neighbors of node i. & Network data. \\ \hline
        Average clustering & Compute the average clustering coefficient for the graph G. The clustering coefficient for the graph is the average, where is the number of nodes in G. & Network data. \\ \hline
\end{longtable}

\subsection{Prediction Results}

\renewcommand{\thefigure}{S\arabic{figure}}
\label{Partial_Features_MAE}
\begin{figure*}[ht]
\centering
    \centering
    \includegraphics[scale=0.65]{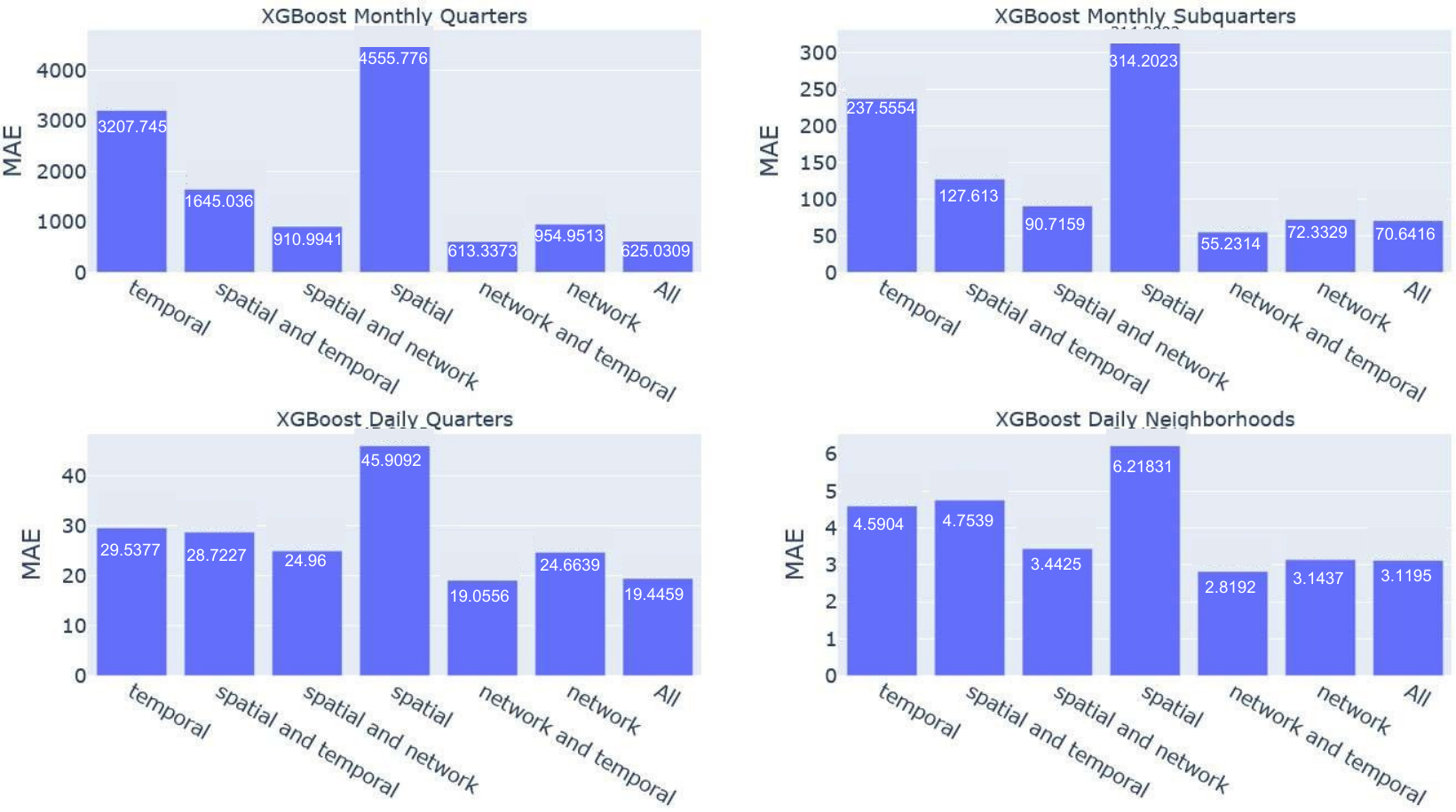}
    \caption{XGBoost Partial Features Groups MAE Comparison}
    \label{fig60:view}
\end{figure*}

\renewcommand{\thefigure}{S\arabic{figure}}
\begin{figure*}[ht]
\centering
    \centering
    \includegraphics[scale=0.6]{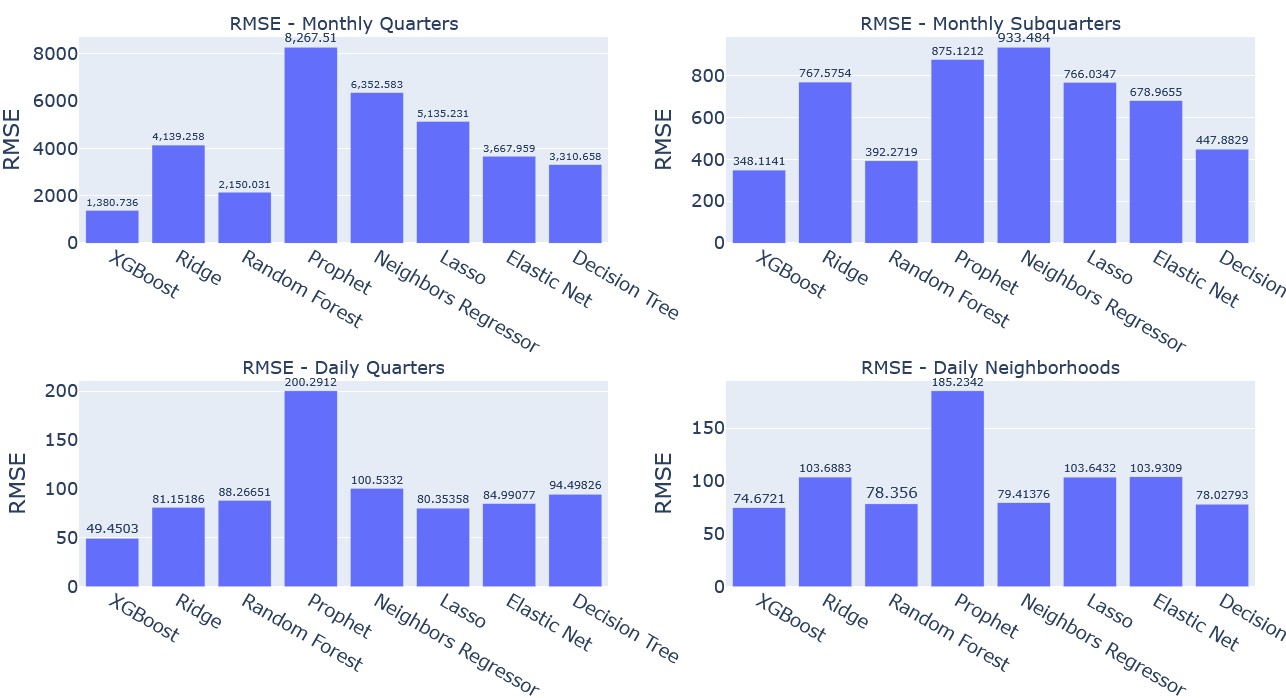}
    \caption{Comparison of RMSE for various ML models, time frames and spatial divisions}
    \label{fig61:view}
\end{figure*}

\renewcommand{\thefigure}{S\arabic{figure}}
\label{Spatial_Analysis}
\begin{figure*}[ht]
    \centering    
    \begin{subfigure}{0.3\textwidth}
        \centering
        \includegraphics[scale=0.25]{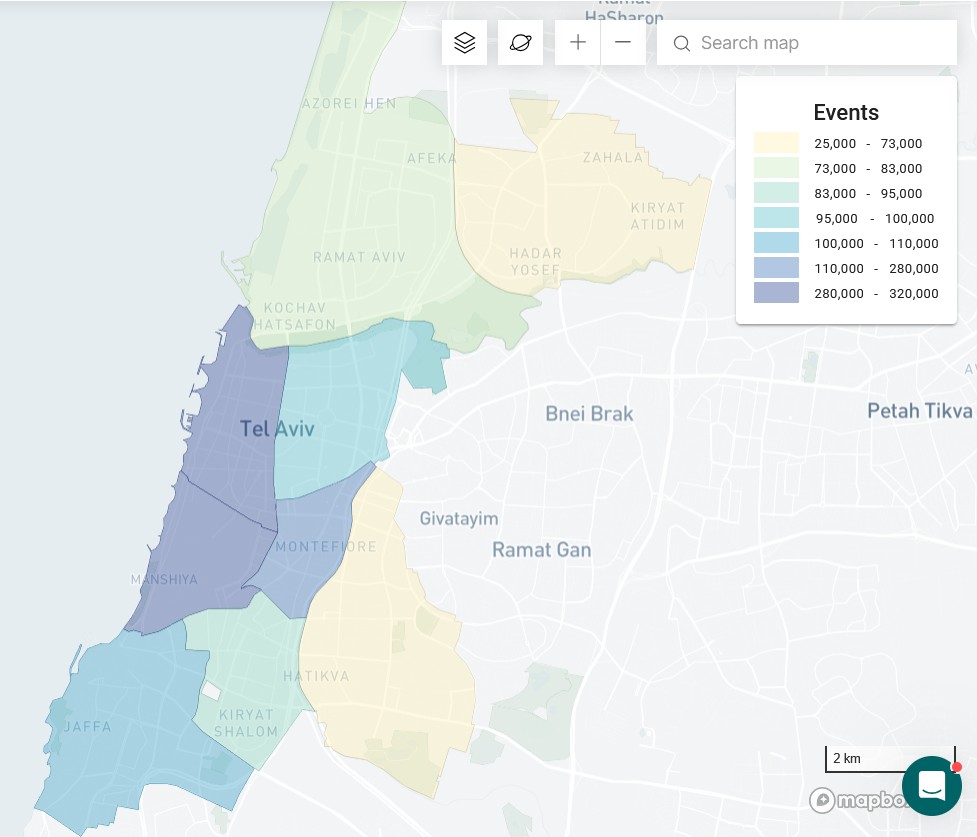}
    \end{subfigure}%
    \hfill
    \begin{subfigure}{0.3\textwidth}
    \includegraphics[scale=0.25]{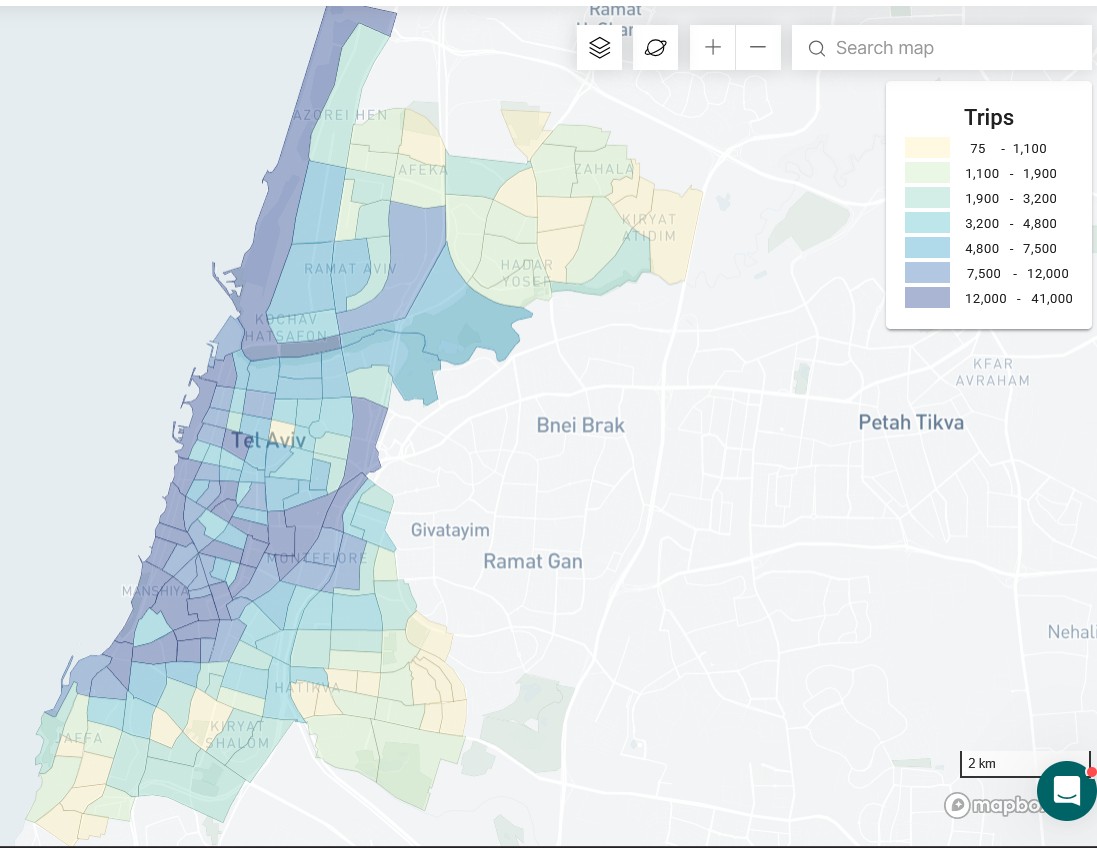}
    \end{subfigure}
    \hfill
    \includegraphics[scale=0.25]{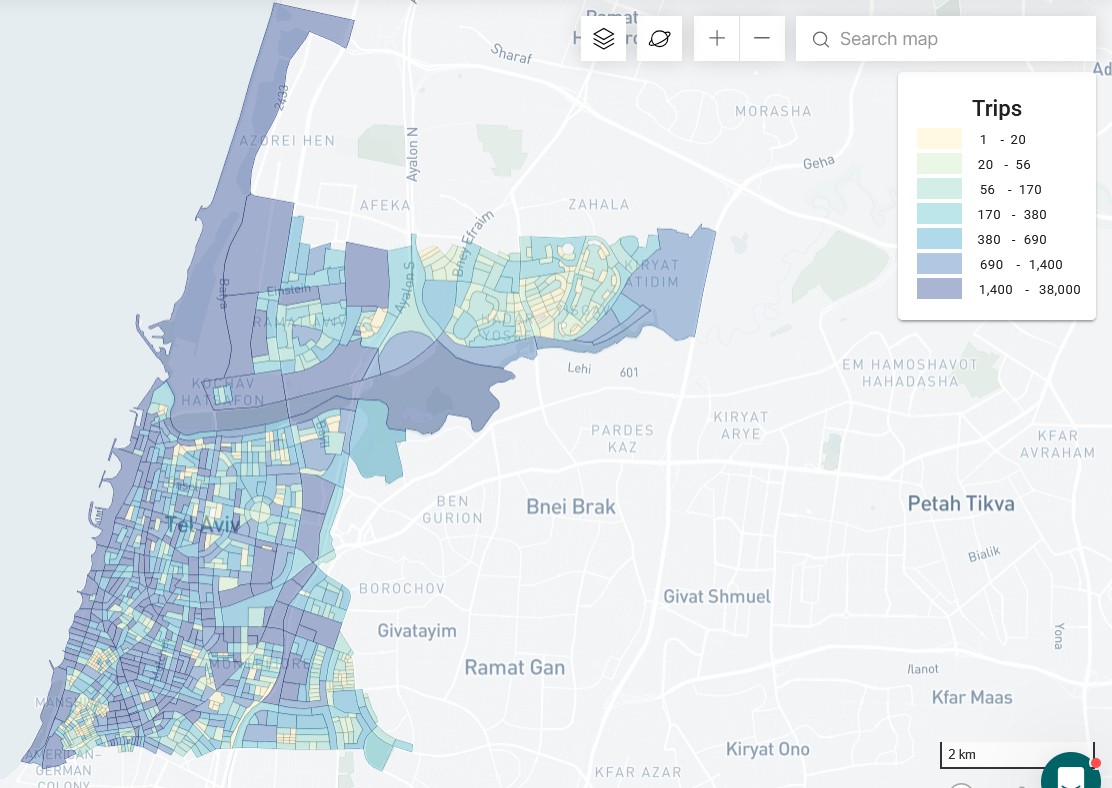}

    \caption{Shared Micromobility Trip Origins in June 2022 By Quarters, Statistical Areas and City Blocks}
    \label{fig:combined}
\end{figure*}

\begin{figure*}[ht]
\centering
    \centering
    \includegraphics[scale=0.6]{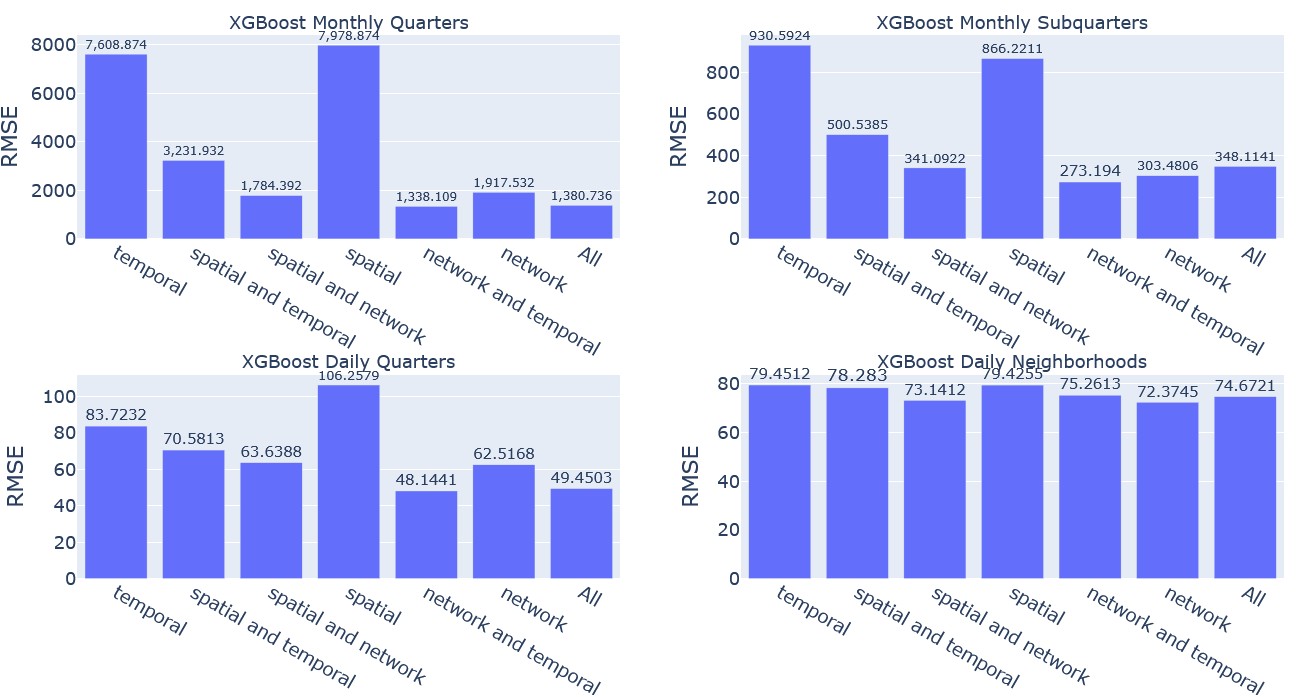}
    \caption{XGBoost Partial Features Groups MAE Comparison}
    \label{fig71:view}
    \label{Partial_Features_RMSE}
\end{figure*}

\renewcommand{\thefigure}{S\arabic{figure}}
\begin{figure*}[ht]
    \centering    
    \begin{subfigure}{0.48\textwidth}
        \centering
\includegraphics[scale=0.5]        {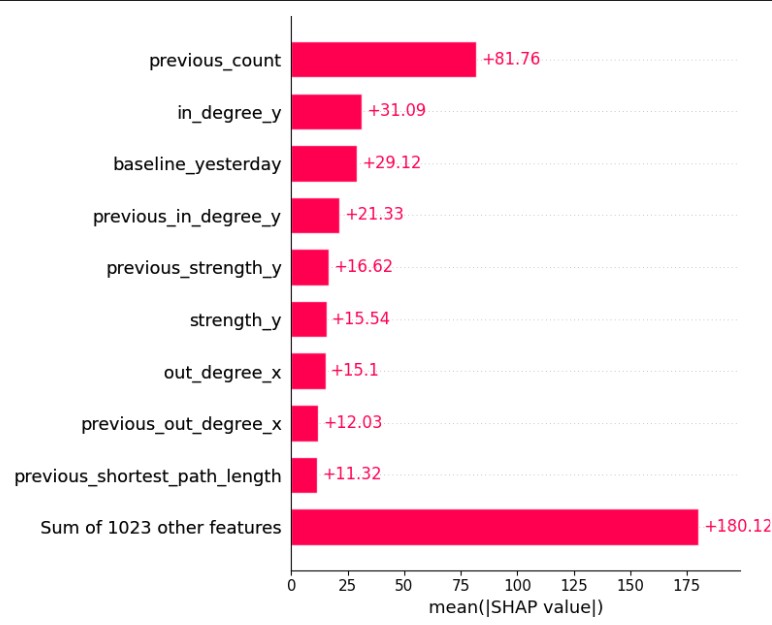}
    \label{fig19:view}
    \end{subfigure}%
    \hfill
    \begin{subfigure}{0.48\textwidth}
    \includegraphics[scale=0.5]{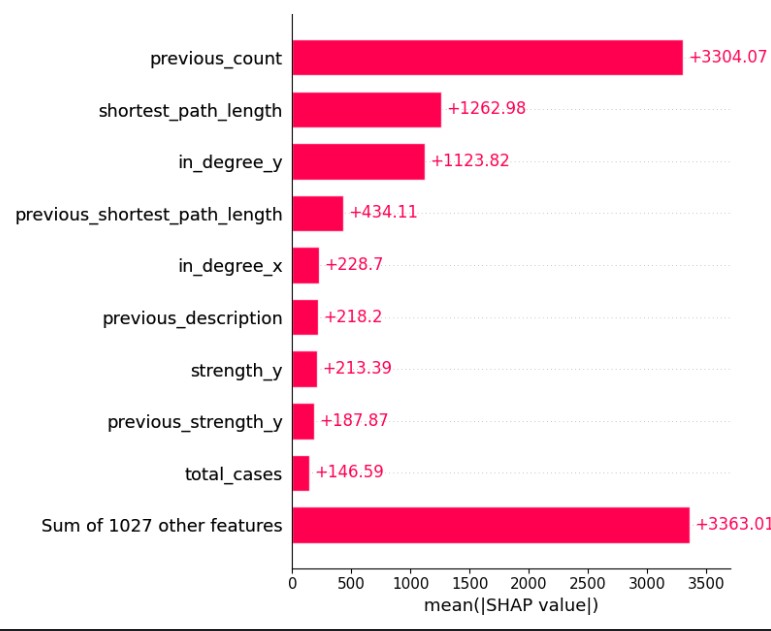}
    \label{fig20:view}
    \end{subfigure}

    \begin{subfigure}{0.48\textwidth}
        \centering
\includegraphics[scale=0.5]                {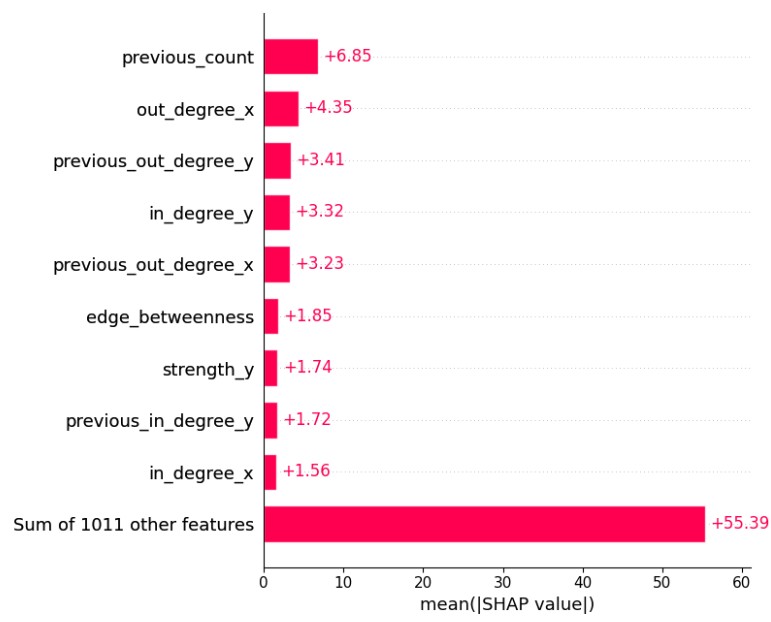}
    \label{fig8:view}
    \end{subfigure}%
    \hfill
    \begin{subfigure}{0.48\textwidth}
    \includegraphics[scale=0.5]{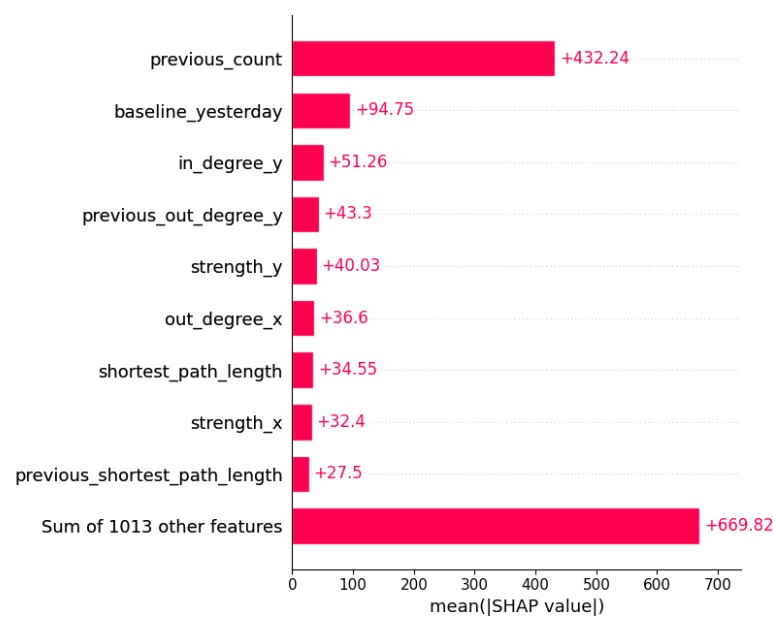}
    \label{fig9:view}
    \end{subfigure}
    \hfill

    \begin{subfigure}{0.48\textwidth}
        \centering
\includegraphics[scale=0.5]                {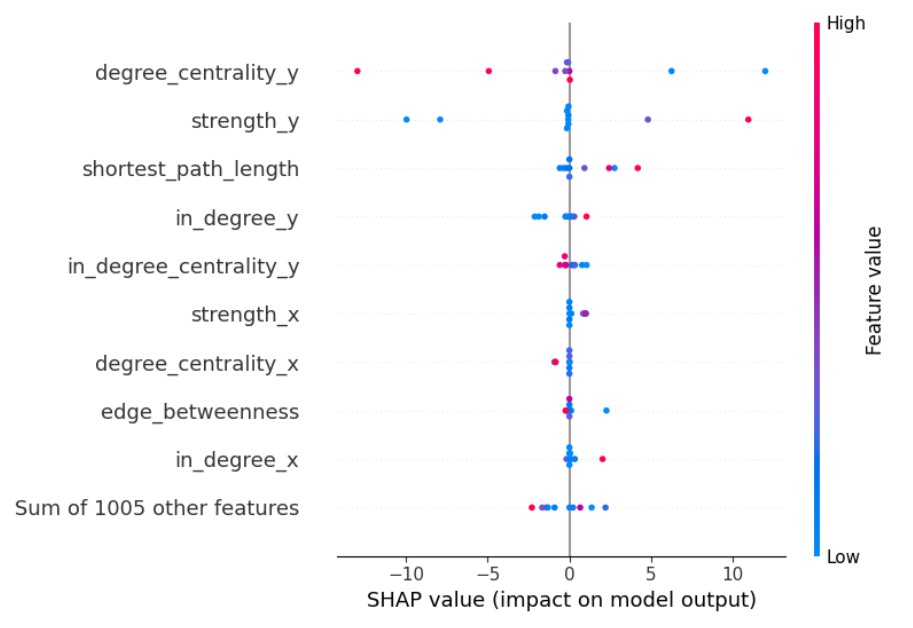}
    \label{fig10:view}
    \end{subfigure}%
    \hfill
    \begin{subfigure}{0.48\textwidth}
    \includegraphics[scale=0.5]{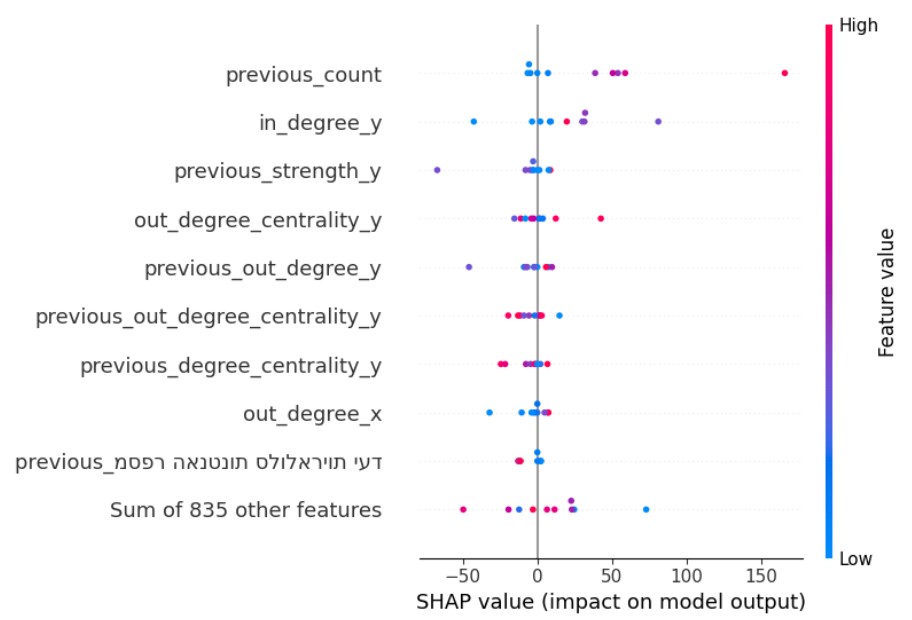}
    \label{fig11:view}
    \end{subfigure}
    \hfill
    \caption{SHAP 10 Features Analysis Over Neighborhoods, Subquarters, Quarters over Monthly and Daily Time Frames}

    \label{SHAP_Analysis}
    \label{fig:combined1}
\end{figure*}

\label{Appendix_Prediction_Results}
\renewcommand{\thetable}{S\arabic{table}}
\onecolumn
\begin{longtable}{|p{0.08\textwidth}|p{0.08\textwidth}|p{0.10\textwidth}|p{0.10\textwidth}|p{0.09\textwidth}|p{0.09\textwidth}|p{0.09\textwidth}|p{0.09\textwidth}|p{0.09\textwidth}|}
\caption{Table of the results of our models over all features:}: \label{Table S6} \\
\toprule
$timeframe$ & $geography$ & $featurestypes$ & $regressortype$ & $regressor$ & $mae$ & $mape$ & $mse$ & $rmse$ \\
\midrule
\endfirsthead
\toprule
$timeframe$ & $geography$ & $featurestypes$ & $regressortype$ & $regressor$ & $mae$ & $mape$ & $mse$ & $rmse$ \\
\midrule
\endhead
\midrule
\multicolumn{9}{|r|}{Continued on next page} \\
\midrule
\endfoot
\bottomrule
\endlastfoot
\multirow{8}{4em}{monthly} & \multirow{8}{4em}{Quarters} & \multirow{8}{4em}{All} & \multirow{8}{4em}{Classical
ML} & Linear Regression & 1.580e+12 & 1.705e+10& 4.510e+24 & 2.120e+12 \\
& & & &  Ridge & 2918.95 & 18.97 & 1.713e+07 & 4139.26 \\
& & & & Lasso & 3906.56 & 25.36 & 2.637e+07 & 5135.23 \\
& & & & Elastic Net & 2549.8 & 14.12 & 1.345e+07 & 3667.96 \\
& & & & Decision Tree & 1693.92 & 6.58 & 1.096e+07 & 3310.66 \\
& & & & Random Forest & 1240.61 & 4.75 & 4.623e+06 & 2150.03 \\
& & & & \textbf{XGBoost} & \textbf{625.03} & \textbf{0.598} & \textbf{1.906e+06} & \textbf{1380.74} \\
& & & & Neighbors Regressor & 3392.83 & 6.36 & 4.036e+07 & 6352.58 \\
& & & & Prophet & 7192.5 & 42.8 & 2.733e+08 & 8267.51 \\
\hline

\multirow{8}{4em}{monthly} & \multirow{8}{4em}{Subquarters} & \multirow{8}{4em}{All} & \multirow{8}{4em}{Classical
ML} & Linear Regression & 5.250e+11 & 6.176e+10& 4.510e+23 & 6.720e+11 \\
& & &  & Ridge & 442.84 & 47.39 & 5.892e+05 & 767.58 \\
& & &  & Lasso & 432.89 & 44.97 & 5.868e+05 & 766.03 \\
& & &  & Elastic Net & 356.83 & 32.09 & 4.610e+05 & 678.97 \\
& & &  & Decision Tree & 109.02 & 3.83 & 2.006e+05 & 447.88 \\
& & &  & Random Forest & 103.54 & 4 & 1.539e+05 & 392.27 \\
& & &  & \textbf{XGBoost} & \textbf{70.6416} & 3.98 & \textbf{1.21e+05} & \textbf{348.11} \\
& & &  & Neighbors Regressor & 252.27 & 7.95 & 8.714e+05 & 933.48 \\  
& & & & Prophet & 97.41 & 35.1 & 4.897e+06 & 875.12 \\
\hline
\multirow{8}{4em}{monthly} & \multirow{8}{4em}{Neighbor-hoods} & \multirow{8}{4em}{All} & \multirow{8}{4em}{Classical
ML} & Linear Regression & 4.600e+11 & 1.490e+11 & 3.970e+23 & 6.300e+11 \\
& & & & Ridge & 164 & 38.4 & 2.105e+05 & 458.77 \\
& & & & Lasso & 154.29 & 34.87 & 2.081e+05 & 456.14 \\
& & & & Elastic Net & 118.64 & 24.96 & 1.561e+05 & 395.08 \\
& & & & Decision Tree & 52.96 & 6.44 & 1.226e+05 & 350.1 \\
& & & & Random Forest & 47.68 & 6.36 & 1.008e+05 & 317.44 \\
& & & & \textbf{XGBoost} & \textbf{24.92} & \textbf{1.44} & \textbf{6.601e+04} & \textbf{256.92} \\
& & & & Neighbors Regressor & 68.41 & 6.51 & 1.986e+05 & 445.64\\ 
& & & & Prophet & 164.004 & 26.54 & 3.481e+05 & 590.117\\
\hline
\multirow{8}{4em}{monthly} & \multirow{8}{4em}{Statistical Areas} & \multirow{8}{4em}{All} & \multirow{8}{4em}{Classical
ML} & Linear Regression & 1764.6 & 1753.39 & 6.311e+09 & 7.944e+04 \\
& & & & Ridge & 13.52 & 2.29 & 1363.51 & 36.93 \\
& & & & Lasso & 11.09 & 1.85 & 1299.41 & 36.05 \\
& & & & Elastic Net & 13.91 & 1.94 & 3341.84 & 57.81 \\
& & & & Decision Tree & 8.13 & 0.6 & 1109.27 & 33.31 \\
& & & & Random Forest & 6.77 & 0.57 & 487.16 & 22.07 \\
& & & & \textbf{XGBoost} & \textbf{3.486} & \textbf{0.247} & \textbf{422.17} & \textbf{20.55} \\
& & & & Neighbors Regressor & 14.44 & 1.27 & 6009.53 & 77.52 \\
& & &  & \\ & & & & Prophet & 35.8 & 12.32 & 9543.562 & 97.69 \\
\hline
\multirow{8}{4em}{daily} & \multirow{8}{4em}{Quarters} & \multirow{8}{4em}{All} & \multirow{8}{4em}{Classical
ML} & Linear Regression & 2.704e+10 & 3.463e+09 & 1.300e+21 & 3.602e+10 \\
& & & & Ridge & 46.08 & 3.25 & 6585.62 & 81.15 \\
& & & & Lasso & 43.77 & 2.84 & 6456.7 & 80.35 \\
& & & & Elastic Net & 36.77 & 1.34 & 7223.43 & 84.99 \\
& & & & Decision Tree & 40.51 & 1.16 & 8929.92 & 94.5 \\
& & & & Random Forest & 37.28 & 1.14 & 7790.98 & 88.27 \\
& & & & \textbf{XGBoost} & \textbf{19.44} & \textbf{0.507} & \textbf{2445.33} & \textbf{49.45} \\
& & & & Neighbors Regressor & 39.78 & 0.72 & 1.011e+04 & 100.53
\\& & &  & Prophet & 180.082 & 21.11 & 102079.31 & 200.29 \\
\hline
\multirow{8}{4em}{daily} & \multirow{8}{4em}{Subquarters} & \multirow{8}{4em}{All} & \multirow{8}{4em}{Classical
ML} & Linear Regression & 1.125e+08 & 4.673e+07 & 2.150e+16 & 1.467e+08 \\
& & & & Ridge & 9.11 & 2.65 & 451.79 & 21.26 \\
& & & & Lasso & 7.63 & 1.86 & 441.69 & 21.02 \\
& & & & Elastic Net & 6.83 & 1.36 & 459.09 & 21.43 \\
& & & & Decision Tree & 5.94 & 0.56 & 584.97 & 24.19 \\
& & & & Random Forest & 5.64 & \textbf{0.53} & 468.61 & 21.65 \\
& & & & \textbf{XGBoost} & \textbf{5.16} & 0.71 & \textbf{413.57} & \textbf{20.34}\\
& & & & Prophet & 19.55 & 8.77 & 1840.966 & 22.47\\
\hline
\multirow{8}{4em}{daily} & \multirow{8}{4em}{Neighbor-hoods} & \multirow{8}{4em}{All} & \multirow{8}{4em}{Classical
ML} & Linear Regression & 1.250e+06 & 6.973e+05 & 2.940e+12 & 1.715e+06 \\
& & & & Ridge & 8.03 & 3.51 & 1.075e+04 & 103.69 \\
& & & & Lasso & 6.21 & 2.33 & 1.074e+04 & 103.64 \\
& & & & Elastic Net & 5.39 & 1.91 & 1.080e+04 & 103.93 \\
& & & & Decision Tree & 4.46 & 0.91 & 6088.36 & 78.03 \\
& & & & Random Forest & 4.26 & 0.92 & 6139.66 & 78.36 \\
& & & & \textbf{XGBoost} & \textbf{3.12} & \textbf{0.624} & \textbf{5575.92} & \textbf{74.67} \\
& & & & Neighbors Regressor & 4.52 & 1.02 & 6306.55 & 79.41 \\ & & & & Prophet & 4.52 & 1.02 & 6306.55 & 79.41
\end{longtable}

\subsection{XGBoost Partial Feature Groups Results}
\label{Appendix_Partial_Feature_Groups}
\renewcommand{\thetable}{S\arabic{table}}
\onecolumn
\begin{longtable}{|p{0.08\textwidth}|p{0.08\textwidth}|p{0.10\textwidth}|p{0.10\textwidth}|p{0.09\textwidth}|p{0.09\textwidth}|p{0.09\textwidth}|p{0.09\textwidth}|p{0.09\textwidth}|}
\caption{Table of the results of partial feature groups over best-performing model (XGBoost):}: \label{Table S7} \\
\toprule
$timeframe$ & $geography$ & $featurestypes$ & $regressortype$ & $regressor$ & $mae$ & $mape$ & $mse$ & $rmse$ \\
\midrule
\endfirsthead
\toprule
$timeframe$ & $geography$ & $featurestypes$ & $regressortype$ & $regressor$ & $mae$ & $mape$ & $mse$ & $rmse$ \\
\midrule
\endhead
\midrule
\multicolumn{9}{|r|}{Continued on next page} \\
\midrule
\endfoot
\bottomrule
\endlastfoot
\multirow{7}{4em}{monthly} & \multirow{7}{4em}{Quarters}
& All & Classical ML & XGBoost & 625.0309 & 0.5967 & 1906432.09 & 1380.7361 \\
        &  & spatial & Classical ML & XGBoost & 4555.7764 & 5.5012 & 63662425.88 & 7978.8737 \\
        & & temporal & Classical ML & XGBoost & 3207.7452 & 3.6127 & 57894962.36 & 7608.8739 \\
        & & network & Classical ML & XGBoost & 954.9513 & 1.0553 & 3676929.703 & 1917.5322 \\
        & & spatial and temporal & Classical ML & XGBoost & 1645.0361 & 2.0752 & 10445381.18 & 3231.9315 \\
        & & network and temporal & Classical ML & XGBoost & 613.3373 & 0.6873 & 1790536.491 & 1338.1093 \\
        & & spatial and network & Classical ML & XGBoost & 910.9941 & 0.9008 & 3184056.115 & 1784.3924 \\
\hline
\multirow{7}{4em}{monthly} & \multirow{7}{4em}{Subquarters} 
& All & Classical ML & XGBoost & 70.6416 & 3.98 & 121183.3971 & 348.1141 \\
        & & spatial & Classical ML & XGBoost & 314.2023 & 14.8464 & 750338.9533 & 866.2211 \\
        & & temporal & Classical ML & XGBoost & 237.5554 & 8.3583 & 866002.2617 & 930.5924 \\
        & & network & Classical ML & XGBoost & 72.3329 & 2.2271 & 92100.4992 & 303.4806 \\
        & & spatial and temporal & Classical ML & XGBoost & 127.613 & 5.5736 & 250538.7936 & 500.5385 \\
        & & network and temporal & Classical ML & XGBoost & 55.2314 & 1.5551 & 74634.9678 & 273.194 \\
        & & spatial and network & Classical ML & XGBoost & 90.7159 & 3.7254 & 116343.9054 & 341.0922 \\
\hline
\multirow{7}{4em}{monthly} & \multirow{7}{4em}{Statistical Areas} 
& All & Classical ML & XGBoost & 3.486 & 0.2469 & 422.1716 & 20.5468 \\
        & & spatial & Classical ML & XGBoost & 24.3498 & 2.4042 & 9730.4518 & 98.6431 \\
        & & temporal & Classical ML & XGBoost & 16.7209 & 2.1285 & 7109.0813 & 84.3154 \\
        & & network & Classical ML & XGBoost & 4.1727 & 0.249 & 416.2324 & 20.4018 \\
        & & spatial and temporal & Classical ML & XGBoost & 9.6455 & 1.367 & 563.0135 & 23.7279 \\
        & & network and temporal & Classical ML & XGBoost & 3.3511 & 0.1664 & 349.6717 & 18.6995 \\
        & & spatial and network & Classical ML & XGBoost & 4.6 & 0.3619 & 506.6839 & 22.5096 \\
\hline
\multirow{7}{4em}{daily} & \multirow{4}{4em}{Quarters} & 
 All & Classical ML & XGBoost & 19.4459 & 0.5078 & 2445.331 & 49.4503 \\
        & & spatial & Classical ML & XGBoost & 45.9092 & 1.0286 & 11290.7471 & 106.2579 \\
        & & temporal & Classical ML & XGBoost & 29.5377 & 0.6006 & 7009.5826 & 83.7232 \\
        & & network & Classical ML & XGBoost & 24.6639 & 0.5146 & 3908.3447 & 62.5168 \\
        & & spatial and temporal & Classical ML & XGBoost & 28.7227 & 0.8507 & 4981.7259 & 70.5813 \\
        & & network and temporal & Classical ML & XGBoost & 19.0556 & 0.4705 & 2317.8549 & 48.1441 \\
        & & spatial and network & Classical ML & XGBoost & 24.96 & 0.5635 & 4049.8929 & 63.6388 \\
\hline
daily & Subquarters & All & Classical ML & XGBoost & 5.16 & 0.71 & 413.57 & 20.34 \\
\hline
\multirow{7}{4em}{daily} & \multirow{7}{4em}{Neighbor-hoods}

& All & Classical ML & XGBoost & 3.1195 & 0.624 & 5575.9221 & 74.6721 \\
        & & spatial & Classical ML & XGBoost & 6.218 & 1.8231 & 6308.4083 & 79.4255 \\
        & & temporal & Classical ML & XGBoost & 4.5904 & 1.2039 & 6312.4933 & 79.4512 \\
        & & network & Classical ML & XGBoost & 3.1437 & 0.5348 & 5238.0721 & 72.3745 \\
        & & spatial and temporal & Classical ML & XGBoost & 4.7539 & 1.4941 & 6128.2315 & 78.283 \\
        & & network and temporal & Classical ML & XGBoost & 2.8192 & 0.5038 & 5664.2666 & 75.2613 \\
        & & spatial and network & Classical ML & XGBoost & 3.4425 & 0.7006 & 5349.6342 & 73.1412 \\
\end{longtable}


\end{document}